\documentclass[letterpaper]{article} % DO NOT CHANGE THIS
\usepackage{aaai2026}  % DO NOT CHANGE THIS
\usepackage{times}  % DO NOT CHANGE THIS
\usepackage{helvet}  % DO NOT CHANGE THIS
\usepackage{courier}  % DO NOT CHANGE THIS
\usepackage[hyphens]{url}  % DO NOT CHANGE THIS
\usepackage{graphicx} % DO NOT CHANGE THIS
\usepackage{natbib}  % DO NOT CHANGE THIS AND DO NOT ADD ANY OPTIONS TO IT
\usepackage{caption} % DO NOT CHANGE THIS AND DO NOT ADD ANY OPTIONS TO IT
\usepackage{algorithm}

\usepackage{amsmath}
\usepackage{xcolor}

\usepackage{booktabs}

\usepackage{multirow}
\usepackage{graphicx}
\usepackage{colortbl}
\usepackage{subfigure}

\usepackage{algpseudocode}

\usepackage{amssymb}
\usepackage{amsmath}

\usepackage{newfloat}
\usepackage{listings}
\DeclareCaptionStyle{ruled}{labelfont=normalfont,labelsep=colon,strut=off} % DO NOT CHANGE THIS
\floatstyle{ruled}
\newfloat{listing}{tb}{lst}{}
\floatname{listing}{Listing}
\newcommand{\nosection}[1]{\vspace{2pt}\noindent\textbf{#1.}}

\newcommand{\modelname}{\texttt{CREAT}}

\newcommand{\ie}{\emph{i.e., }}
\newcommand{\eg}{\emph{e.g., }}

\title{Potent but Stealthy: Rethink Profile Pollution against Sequential \\ Recommendation via Bi-level Constrained Reinforcement Paradigm}
\author{
    Jiajie Su\textsuperscript{\rm1}\equalcontrib,
    Zihan Nan\textsuperscript{\rm2}\equalcontrib,
    Yunshan Ma\textsuperscript{\rm3},
    Xiaobo Xia\textsuperscript{\rm4,5},
    Xiaohua Feng\textsuperscript{\rm1},\\
    Weiming Liu\textsuperscript{\rm6},
    Xiang Chen\textsuperscript{\rm1},
    Xiaolin Zheng\textsuperscript{\rm1}\thanks{Xiaolin Zheng is the corresponding author.},
    Chaochao Chen\textsuperscript{\rm1}
}
\affiliations{
    \textsuperscript{\rm1} Zhejiang University, 
    \textsuperscript{\rm2} Peking University\\
    \textsuperscript{\rm3} Singapore Management University,  
    \textsuperscript{\rm4} National University of Singapore\\
    \textsuperscript{\rm5} MoE Key Laboratory of Brain-inspired Intelligent Perception and Cognition,\\ University of Science and Technology of China\\
    \textsuperscript{\rm6} Tiktok, Bytedance\\
    \{sujiajie, fengxiaohua, wasdnsxchen, xlzheng, zjuccc\}@zju.edu.cn, \\ zhnan25@stu.pku.edu.cn, ysma@smu.edu.sg, xbx@nus.edu.sg, lwming95@gmail.com
}

\begin{document}

\maketitle

\begin{abstract}
Sequential Recommenders, which exploit dynamic user intents through interaction sequences, is vulnerable to adversarial attacks.
While existing attacks primarily rely on data poisoning, they require large-scale user access or fake profiles thus lacking practicality.
In this paper, we focus on the Profile Pollution Attack that subtly contaminates partial user interactions to induce targeted mispredictions.
Previous PPA methods suffer from two limitations, \ie i) over-reliance on sequence horizon impact restricts fine-grained perturbations on item transitions, and ii) holistic modifications cause detectable distribution shifts.
To address these challenges, we propose a constrained reinforcement driven attack \modelname~that synergizes a bi-level optimization framework with multi-reward reinforcement learning to balance adversarial efficacy and stealthiness.
We first develop a Pattern Balanced Rewarding Policy, which integrates pattern inversion rewards to invert critical patterns and distribution consistency rewards to minimize detectable shifts via unbalanced co-optimal transport.
Then we employ a Constrained Group Relative Reinforcement Learning paradigm, enabling step-wise perturbations through dynamic barrier constraints and group-shared experience replay, achieving targeted pollution with minimal detectability.
Extensive experiments demonstrate the effectiveness of \modelname.
The implementation code is released at \url{https://github.com/SSndot/CREAT}.
\end{abstract}

% Uncomment the following to link to your code, datasets, an extended version or similar.
% You must keep this block between (not within) the abstract and the main body of the paper.
% \begin{links}
%     \link{Code\&Datasets}{https://aaai.org/example/code}
% \end{links}

\section{Introduction}

%% 1. Introduce profile pollution attack 

%%% What is the sequential recommendation?
Sequential Recommendation (SR) \cite{xie2022contrastive,liu2023joint,su2023behavior} explores user evolving interests to make the next-item prediction.
% why SR is vulnerable to attack
% SR prediction heavily relies on interaction transitions \cite{wang2023poisoning,du2024parl}, thus even slight perturbations of training sequences can lead to substantial shifts in recommender cognition.
% It is vulnerable to attacks.
Although SRs are widely regarded as delivering trustworthy results, their sensitivity to sequential patterns renders them vulnerable to adversarial attacks \cite{survey_poisonrec,du2024parl}.
%%% In this paper, we focus on profile pollution attack
Recent research \cite{wang2023poisoning,zhang2024lorec} has mostly focus on \textit{data poisoning} attacks, which manipulate SR by injecting a substantial amount of crafted sequences.
%%% 
But this attack relies on large-scale access to user accounts or the ability to create numerous fake profiles, which can be impractical in real-world scenarios.
In this paper, we focus on a more targeted and stealthier attack strategy, \textit{Profile Pollution Attack (PPA)}, which subtly contaminates partial individual user interaction histories to corrupt SR into a targeted misprediction on specific subtasks.
%

%% black-box - gradient
%% defending
%% influence
%% 以往的工作 1、没有充分挖掘sequential pattern来最大化modification的impact 2、把这个问题刻画成一个单向优化的问题，缺乏对于隐蔽性的考虑。
% The key challenge of targeted profile pollution lies in how to maximize the attacking impact while introducing minimal modifications to a limited set of user behavioral sequences.
%
%% 2. motivation, previous work all overlook how to maximize the attack intensity while preserving the stealthiness [this is an unexplored question]

\begin{figure}[t]
\centering
\includegraphics[width=1.0\linewidth]{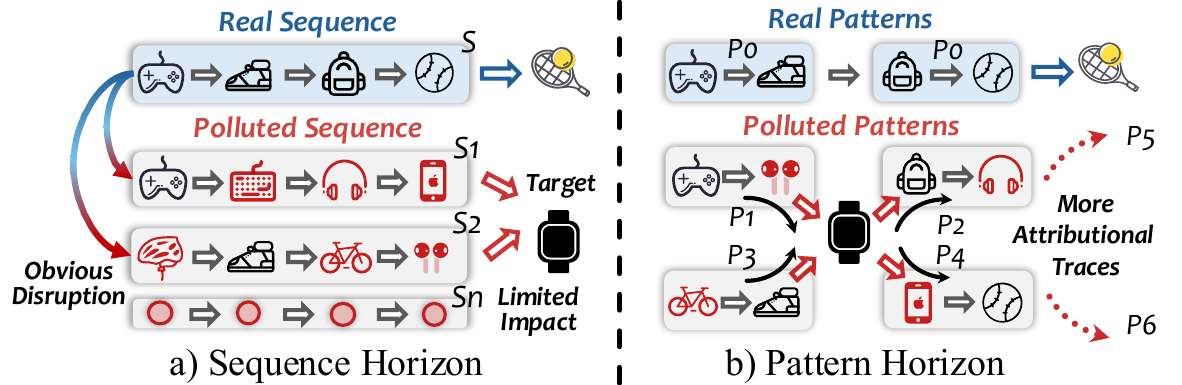}
\vspace{-0.2 in}
\caption{The motivation of \modelname.} 
\label{story}
\vspace{-0.2 in}
\end{figure}

Several previous studies attempted to conduct PPA against SR.
One line of methods \cite{blackbox,defending} crafts perturbations guided by gradient estimation on the attack loss then inserts influential items.
%
% These gradient-based attacks are constrained by insufficient optimization due to single-step gradient descent \cite{madry2017towards}.
%
Another line \cite{influence} leverages the influence function to measure the modification impact to training sequences on model parameters.
%
% However, the gradient-based attacks are constrained by insufficient optimization due to single-step gradient descent \cite{madry2017towards}, while the influence-based methods tend to yield inaccurate estimations of sample importance in complex backbones due to cumulative reduction operations.
% 他们利用序列形成后的整体排列状态来逆转推荐系统对于物品转换关系的理解。
As Fig \ref{story} shows, these works largely follow a common paradigm, \ie assess the intensity of posterior attacks from the \textit{sequence horizon}, which leverages the global structure of polluted sequences to modify recommender comprehension on user behaviors.
%
% measuring and maximizing the attack strength at the polluted \textit{sequence-level}.
%
Such a paradigm faces two limitations,
\textbf{(i) Low attack intensity}. Relying on amplification of the whole polluted sequence effect confines the attack to user subsets with specific interests, failing to reshape model perception on fine-grained sequential transitions.
\textbf{(ii) Subtle attack stealth}. Forcing the overall interests of sequences to align with attack targets induces a noticeable distributional shift. Since interference at the sequence horizon is a coarse-grained and user-level disruption, it often requires manipulating numerous sequences to achieve a significant effect, thus increasing the detection risk.
%
% Forcing overall interests of sequences to align with attack targets induces a noticeable distributional shift. Given the limited number of sequences available for manipulation, this shift substantially raises the risk of being detected.
%
These issues motivate us to a crucial question:
\textbf{\textit{How to exploit pivotal structures underlying the recommender collaborative modeling to maximize attack strength with minimal detectability?}}

%% 我们将ppa问题重构成一个双层优化的问题，解释如何bi-level
In light of this, we reformulate PPA into a bi-level optimization problem, wherein the \textit{upper-level objective} seeks to maximize the utility of sequence perturbations, subject to a \textit{lower-level constraint} enforcing a bounded degree of stealthiness in the crafted sequences.
% bi-level的2个level各自会遇到的问题，以及协同优化会遇到的问题
The upper-level formulation is grounded in a new theoretical concept, termed \textit{pattern horizon}, which postulates that the model's prediction of the next item is inherently driven by an attributional trace over sequential pattern dependencies \cite{yin2024dataset,dang2025data}.
As Fig \ref{story} shows, inverting pivotal sequential patterns toward a target item enables a finer-grained exploration of how distinct perturbations propagate through the collaborative reasoning process of SR.
Due to synergistic and cascading effects between patterns \cite{liu2023enhancing}, modifying a subset of patterns can be generalized to multiple similar patterns during the SR model’s learning process, thereby amplifying the adversarial impact.
At the lower level, in contrast to traditional sequence horizon perturbations that operate on holistic representations, the pattern-horizon-guided perturbations emphasize localized structural shifts.
This allows the attacker to strategically calibrate pattern granularity and compositional balance, effectively regulating the distributional deviation between crafted and benign sequences.
However, this bi-level formulation raises three key challenges, \ie 
\textbf{\textit{Ch1: How to discern and reverse critical sequential patterns?}}
\textbf{\textit{Ch2: How to modulate the stealthiness of pattern-balanced perturbations?}}
\textbf{\textit{Ch3: How to synergistically optimize the coupled objectives across both levels?}}

%

%% 4. Overview of our model

To tackle these challenges, we propose a \underline{C}onstrained \underline{RE}inforcement driven \underline{AT}tack, termed \modelname, which leverages a group relative reinforcement learning constrained by stealth-aware conditions for targeted profile pollution.
The key insight of \modelname~lies in simulating the bi-level optimization problem with a multi-reward mechanism, \ie maximize the inversion effect of critical sequential patterns while minimize detection risk, thereby deriving an optimal pollution policy.
% 为了分别实现upper-level和lower-level优化目标控制，我们提出pattern balanced rewarding policy，which 由 pattern inversion reward 和 distribution consistency reward组成。
For separately regulating the upper-level and lower-level objectives, we design a \textit{Pattern Balanced Rewarding Policy (PBRP)}, integrating both inversion and consistency rewards to guide the perturbation.
% 为了挖掘最具影响力的细粒度pattern，我们首先develop the pattern inversion reward
To uncover the most influential patterns (\textbf{\textit{Ch1}}), we first develop the \textit{pattern inversion reward} which identifies the optimal perturbation positions that simultaneously achieve maximal pattern-level semantic inversion and diversification.
To regulate stealthiness of inverted patterns (\textbf{\textit{Ch2}}), the \textit{distribution consistency reward} adapts an unbalanced co-optimal transport to constrain the distributional shifts of polluted sequential representation from both sample and pattern aspects.
Building upon the bi-level mechanism, we establish \textit{Constrained Group Relative Reinforcement Learning (C-GRRL)}, which enables step-wise and self-reflective perturbations over polluted sequences.
This paradigm consists of two stages, \ie critical pattern localization and constrained inversion optimization.
In the \textit{localization stage}, we train a sequence masker solely guided by the inversion reward, aiming to identify positions on target items that yield maximal adversarial impact.
In the \textit{constrained stage}, a dynamic barrier constraint that adaptively joins inversion and consistency rewards fine-tunes the masker, thereby aligning with the dual imperative of maximizing adversarial efficacy while preserving stealthiness (\textbf{\textit{Ch3}}).
Specifically, we employ a constrained group-relative policy within the bi-level optimization, which integrates a group-shared experience replay buffer and relative prioritized sampling, to accelerate the convergence toward optimal multi-step perturbations.

%% 5. contributions

Main contributions are:
(1) We revisit the PPA against SR into a bi-level optimization problem, and propose a novel framework with group relative constrained reinforcement learning.
(2) We devise the PBRP policy, developing pattern inversion reward to extract influential patterns and distribution consistency reward to control pattern stealth.
(3) We establish the C-GRRL paradigm, realizing step-wise and self-reflected perturbation optimization.
(4) Extensive experiments demonstrate the effectiveness of \modelname.

\section{Related Work}

\nosection{Sequential Recommendation}
% 非常简略、但全面地介绍sequential recommendation的发展脉络和所有类型工作（概括式的，不具体距离的）
SR characterizes dynamic user intents by modeling behavioral sequences. 
Early work models sequential patterns with Markov Chain assumption~\cite{rendle2010factorizing}. 
Later, Recurrent Neural Networks \cite{wu2017recurrent}, Convolutional Neural Networks \cite{tang2018personalized}, Graph Neural Networks \cite{wu2019session,zheng2020dgtn,su2023enhancing}, and Transformers \cite{kang2018self} are developed to model interests over interactions.
Besides, unsupervised learning based models \cite{xie2022contrastive} extract more informative user patterns by deriving self-supervision signals.
Inspired by generative models, a series of diffusion-based SRs \cite{yang2023generate} merge, leveraging diffusion generative capabilities to produce personalized content.
A series of methods \cite{li2023gpt4rec,liu2024llm} utilize large language models to enhance the accuracy of SR.
But the vulnerability of SR to adversarial attacks based on malicious sequences remains a significant security problem.

\nosection{Adversarial Attacks in Recommender Systems}
Adversarial attacks \cite{zhang2021data,zhang2022pipattack,wang2023poisoning,wang2024poisoning} on recommender systems can be categorized into: (1) \textit{data poisoning} and (2) \textit{profile pollution}.
%
% 参考版本1: Data poisoning attacks craft fake user profiles and inject such profiles into the training data, causing biased recommendation results upon deployment. Unlike poisoning attacks, profile pollution attacks alter existing user profiles (i.e., user interactions) to manipulate recommended items. Profile pollution can be performed via security breaches like web injection, CSRF attacks or malware.
% 参考版本2: Data poisoning attacks involve generating fake users by forging userprofiles and injecting them into the training data of the recommender system, thereby resulting in biased recommendations expected by attackers. In contrast, profile pollution attacks are more subtle, and attackers interfere with the recommendation results of recommender systems by modifying, tampering, or forging user behavior data.
%
Data poisoning attacks \cite{zhang2020practical,song2020poisonrec,tang2020revisiting,huang2021data,wu2023influence} compromise recommenders by injecting fabricated user profiles, skewing model outputs toward adversarial objectives.
Conversely, profile pollution attacks (PPA) \cite{blackbox,zhang2021attacking,fan2021attacking,lin2022shilling} directly tamper with user interaction records, subtly distorting individual recommendation streams without requiring large-scale data infiltration.
In this paper, we focus on the PPA and aim to manipulate the recommendation with targeted goals.
Existing studies on PPA is divided into four types, which are respectively based on injection \cite{xing2013take,meng2014your,zhang2019understanding}, replacement \cite{defending}, repetition \cite{tang2018personalized}, and expert knowledge \cite{yang2017fake}.
%
% PPA against SR
The exploration of PPA against SR remains at a nascent stage.
One branch of methods \cite{blackbox,defending} generates perturbations by estimating gradients of attack loss, injecting impactful items into the sequences.
SimAlter \cite{blackbox} appends adversarial items by extending the targeted fast gradient sign method from the continuous to discrete item space.
Replace \cite{defending} typically utilizes the loss gradient to guide the selection of injected items.
However, these gradient-based attacks are constrained by insufficient optimization due to single-step gradient descent \cite{madry2017towards}.
Another branch of work, like INFAttack \cite{influence}, employs influence function to quantify how modifications affect the model parameters.
But the influence computation chain introduces substantial complexity while its accuracy deteriorates with deeper backbones.
Although these works promote PPA to some extent, they assess pollution strength from \textit{sequence horizon}, which overlooks exploring fine-grained patterns, thus constraining attack effectiveness and increasing the detection risk.

\section{Methodology}

\subsection{Problem Formulation}
\nosection{Profile Pollution Attack against SR}
Let \( \varPhi_{\theta} \) denote a sequential recommender with parameters \(\theta\), where users \( u \in \mathcal{U} \) and items \( v \in \mathcal{V} \) are represented by chronological interaction sequences \(\mathbf{s}_u = [v_1, \dots, v_L]\).
The recommender is trained on a dataset \( \mathcal{D} = \{\mathbf{s}_u \mid u \in \mathcal{U}\} \) with the next-item prediction loss \( \mathcal{L}(\cdot) \).
The profile pollution attack (PPA) aims to perturb a certain \textit{subset} of training sequences \( \mathcal{S} \subseteq \mathcal{D} \) into \( \mathcal{S}' \) by replacing limited interactions to maximize the recommendation exposure of a target item \( v^* \in \mathcal{V} \).
For each polluted sequence, the amount of perturbations $M$ is bound by \( M \leq K \), where \( K \) is a small constant.
%
% Note that the attacker knows the recommender architecture and training loss function, and the attack is executed \textit{before} model training.
We assume the attacker knows the model architecture and loss function, or can obtain a surrogate model through prior extraction.
This assumption is justified by recent advances in recommendation model extraction \cite{blackbox,Wang_Su_Chen,fewmea}, which demonstrate that black-box recommenders can be reliably approximated with limited or even no user data, resulting in surrogate models with similar hidden representations and output behavior.
Formally, the objective of PPA is to construct perturbed sequences as: 
\begin{equation}
% \small
\begin{aligned}
\label{eq:goal}
    \hat{\theta} &= \arg\min_{\theta} \sum_{ \mathbf{s}_u\in (\mathcal{D} \setminus \mathcal{S}) \bigcup \mathcal{S}'} \mathcal{L}(\mathbf{s}_u; \theta),\\
    \widetilde{\mathcal{S}} &= \arg\max_{\mathcal{S}'} \mathbb{E}_{u \sim \mathcal{U}} \left[ \text{ER}(v^* \mid \varPhi_{\hat{\theta}}(\mathbf{s}_u)) \right].
\end{aligned}
\nonumber
\end{equation}
Here,\(\hat{\theta}\) denotes the recommender parameters after pollution and \( \widetilde{\mathcal{S}} \) indicates the optimal polluted sequences that invert the recommender training to maximize the exposure ratio \(\text{ER}(v^* \mid \varPhi_{\hat{\theta}}(\mathbf{s}_u))\) of \( v^* \) in recommendation lists.
%
% \(\text{ER}(v^* \mid \varPhi_{\hat{\theta}}(\mathbf{s}_u))\) measures the exposure ratio of \( v^* \) for user \( u \).

\nosection{Framework Overview}
% The aim of \modelname~is optimizing the sequence pollution strategy under the guidance of constrained reinforcement learning to promote the exposure ratio of the target item.
%
We present \modelname~in Figure \ref{framework} and its algorithm in Appendix C.
\modelname~consists of three components, \ie \textit{perturbation masker}, \textit{pattern balanced rewarding policy (PBRP)}, and \textit{constrained group relative reinforcement learning (C-GRRL)}.
The perturbation masker produces the step-wise masking strategies and is optimized by the PBRP with bi-level objectives.
PBRP develops pattern inversion reward to maximize attack intensity, and distribution consistency reward for stealthy constraints.
C-GRRL devises a two-stage optimization to solve the bi-level constrained problem.

\begin{figure*}[t]
\centering
\includegraphics[width=1.0\linewidth]{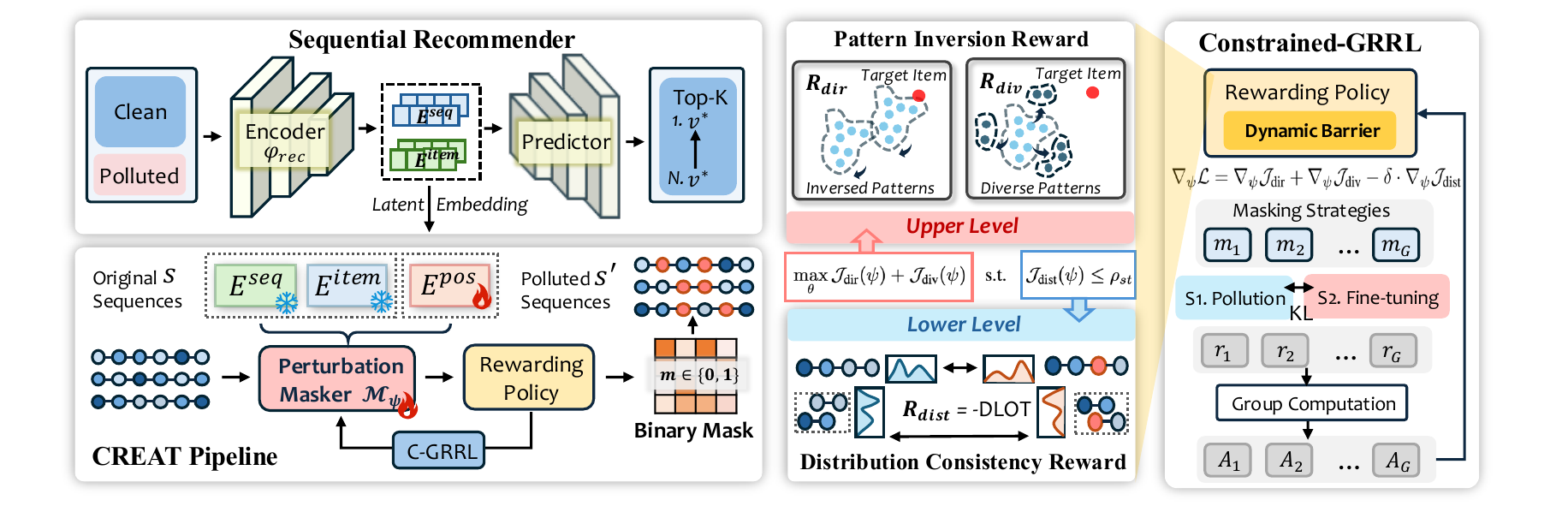}
\vspace{-0.2 in}
\caption{The proposed PPA framework of \modelname, which consists of three components, \ie the perturbation masker, pattern balanced rewarding policy with inversion and consistency reward, and constrained group relative reinforcement learning.} 
\label{framework}
\vspace{-0.2 in}
\end{figure*}

\subsection{Pattern Balanced Rewarding Policy}

\nosection{Perturbation Masker}
% To achieve the PPA objective in Eq.(\ref{eq:goal}), we design a learnable perturbation masker \(\mathcal{M}_{\psi}\), which selectively modify $\mathcal{S}$ by injecting the target item $v^*$ at strategically chosen positions under a perturbation budget \(M \leq K\), thereby biasing the model $\varPhi_{\theta}$ toward favoring $v^*$ in recommendations.
% %
Given a sequential recommender \(\varPhi_{\theta}\) trained on interaction sequences \(\mathcal{D}\), we design a perturbation masker \(\mathcal{M}_{\psi}\) to identify optimal positions in a subset of training sequences \(\mathcal{S} \subseteq \mathcal{D}\) for replacing items with the target item \(v^*\), under a perturbation budget \(M \leq K\).
Formally, for a sequence \(\mathbf{s} = [v_1, \dots, v_L]\), the masker generates a binary mask \(\mathbf{m} \in \{0,1\}^L\) through a \textit{step-wise reinforcement learning process}, where \(\mathbf{m}_t = 1\) indicates replacing \(v_t\) with \(v^*\).
The perturbed sequence is constructed iteratively as:
\begin{equation}
    \mathbf{s}'^{(i)} = \mathbf{s}'^{(i-1)} \odot (1 - \mathbf{m}^{(i)}) + v^* \cdot \mathbf{m}^{(i)},
    \nonumber
\end{equation}
where \(\mathbf{m}^{(i)}\) is the mask vector at step \(i\), and \(\odot\) denotes element-wise multiplication.
%
% The final perturbed sequence after \(M\) steps is \(\mathbf{s}' = \mathbf{s}'^{(M)}\).
%
Unlike traditional perturbation, our masker follows a pattern balanced rewarding policy.
At each step \(i\), the masker selects the next position to perturb based on the current state \(\mathbf{s}'^{(i-1)}\), and receives the reward based on the adversarial impact of the perturbation.
%
% This feedback loop ensures that each perturbation decision is dynamically adjusted based on cumulative rewards from prior steps, enabling adaptive trade-offs between attack efficacy and stealth.

\nosection{Pattern Inversion Reward}
% pattern inversion + pattern diversity
To amplify the attack effect, we propose a \textit{Pattern Inversion Reward} that guides the masker to identify sub-pattern positions whose semantic distributions are notably different from that of the target item. By inserting the target item adjacent to these semantically divergent sub-patterns, the attacker can construct \textit{spurious attributional paths} that link diverse user intent patterns to the target item. This misleads the recommender into falsely associating varied behavioral cues with the target, thereby increasing its exposure.
This reward operates on two complementary dimensions, \ie the directionality and diversity of inversion pathways.
\textbf{For the directionality of inversion}, we encourage the masker to maximize the semantic distance between the target item with both historical and future sequential contexts.
Let \(T^{(i)} = \{t_1, \dots, t_i\}\) denote perturbed positions up to step \(i\).
For each \(t_j \in T^{(i)}\), we compute embeddings of the predecessor \(S^p_{t_j} = \mathbf{s}'^{(i)}_{[1:t_j-1]}\) and successor \(S^f_{t_j} = \mathbf{s}'^{(i)}_{[t_j+1:L]}\) using the representation encoder \(\varphi_{\text{rec}}\) of \(\varPhi_{\theta}\).
The directionality reward at step \(i\) is:
\begin{equation}
\tiny
    R_{\text{dir}}^{(i)} = \sum_{j=1}^i \left[ D\left(\varphi_{\text{rec}}(S^p_{t_j}), \varphi_{\text{rec}}(v^*)\right) + D\left(\varphi_{\text{rec}}(v^*), \varphi_{\text{rec}}(S^f_{t_j})\right) \right],
\nonumber
\end{equation}
where \(D(\cdot, \cdot)\) is the Euclidean distance.
\textbf{For the diversity of inversion}, we refine the strategy by leveraging the synergistic effects among attack patterns.
This inversion reward is designed to enhance the divergence among attack modes, ensuring heterogeneous attack paths and reduce detection risk.
%
% To avoid repetitive attack patterns, we maximize the diversity of perturbed sequences using a \textit{determinantal point process (DPP)}.
%
At step \(i\), let \(\mathcal{Y}^{(i)}\) be the set of subsequences in \(\mathbf{s}'^{(i)}\) that exclude target items, we map each subsequence \(y_p \in \mathcal{Y}^{(i)}\) to a unit-norm prototype as \(\tilde{\varphi}(y_p) = \varphi_{\text{rec}}(y_p) / \|\varphi_{\text{rec}}(y_p)\|\). 
Then we form the diversity reward with the Gram matrix \(G_y^{(i)}\):
\begin{equation}
\small
    G_y^{(i)} = \left[\tilde{\varphi}(y_k)^\top \tilde{\varphi}(y_l)\right]_{k,l=1}^{|\mathcal{Y}^{(i)}|}, \ R_{\text{div}}^{(i)} = \log \det\left(G_y^{(i)}\right).
    \nonumber
\end{equation}
% Maximizing \(\det(G_y^{(i)})\) encourages orthogonal prototypes, ensuring heterogeneous attack paths and reducing detection risk.

\nosection{Distribution Consistency Reward}
% Unbalanced CO-Optimal Transport 其实是 Unbalanced OT 和 CO-OT 的结合：
% 1、Unbalanced OT 和经典OT的区别是，放松了对于迁移质量相等的约束，所以可以更好地处理分布中的一些离群点问题，实现robust OT。（这里和我们的关系是，我们希望这个在隐层的虚假序列对齐和真实序列对齐是鲁棒的，也就是说我们追求fake group的整体隐蔽性，如果有部分fake seq较为离群，不会在这些outlier上产生过多的cost）
% 2、CO-Optimal Transport 和经典OT的区别是，经典OT只做sample到sample的迁移，CO-Optimal Transport 既考虑到sample-sample的迁移，又考虑到两个分布间feature-feature的迁移。feature可以理解为sample分布的转置，也就是每一维度feature在一堆数据上的分布。（这里和我们的关系是，我们不仅希望在instance维度上fake seq具有真实性，而且希望fake seq传达出的seq pattern具有真实性，可以认为seq pattern是seq instance的一种细粒度模式，seq instance是inter-seq，seq pattern是intra-seq）
To ensure the stealthiness of perturbed sequences, we introduce the \textit{Distribution Consistency Reward}, which constrains the deviation between polluted sequences and their original counterparts in both instance-level and pattern-level semantics.
We ground this reward in a dual-level co-optimal transport (\texttt{DLOT}) optimization \cite{tran2023unbalanced}, which provides a principled measure of distributional shifts by simultaneously aligning global sequence and local transitional patterns.
At step \(i\), let \(\mathbf{s}\) denote the original sequence and \(\mathbf{s}^{\prime(i)}\) the perturbed sequence.
For the sequence-level, we obtain representations from the recommender encoder as \(
    \mathbf{h}_{\text{orig}} = \varphi_{\text{rec}}(\mathbf{s})\) and \(  
    \mathbf{h}_{\text{pert}}^{(i)} = \varphi_{\text{rec}}(\mathbf{s}^{\prime(i)})
    \).
For the pattern-level, the sets \( \textbf{p}_{\text{orig}}=
    \{\varphi_{\text{rec}}(\mathbf{s}_{[t:t+k]})\}_{t=1}^{L-k} \) and \( \textbf{p}_{\text{pert}}=
    \{\varphi_{\text{rec}}(\mathbf{s}^{\prime(i)}_{[t:t+k]})\}_{t=1}^{L-k} 
    \) contain $k$-gram pattern embeddings derived from sliding windows over $\mathbf{s}$ and $\mathbf{s}^{\prime(i)}$.
Then we construct the sequence-pattern spaces from the dual-level representations:
\begin{equation}
\small
    \mathbb{X}_{\text{orig}} = (\mathbf{h}_{\text{orig}}, \mathbf{p}_{\text{orig}}, \xi_{\text{orig}}), \quad 
\mathbb{X}_{\text{pert}}^{(i)} = (\mathbf{h}_{\text{pert}}^{(i)}, \mathbf{p}_{\text{pert}}^{(i)}, \xi_{\text{pert}}),
\nonumber
\end{equation}
where $\xi_{\text{orig}}$ and $\xi_{\text{pert}}$ are scalar functions that define the sample-feature interactions.
Unlike traditional balanced OT \cite{cuturi2013sinkhorn,flamary2021pot,liu2023coclusterot} and unbalanced OT \cite{pham2020unbalanced,sejourne2022faster,liu2024user}, we incorporate two transport plans in \texttt{DLOT}, \ie $\pi^s$ aligns entire sequences while $\pi^f$ aligns intra-sequence patterns.
%
% Interaction functions $\xi_{\text{orig}}$ and $\xi_{\text{pert}}^{\prime(i)}$ embed sequences and patterns into a shared representation space, \eg via $\varphi_{\text{rec}}(x^s) \oplus \varphi_{\text{rec}}(x^f)$.
%
The optimization of \texttt{DLOT} is:
\begin{equation}
\small
\begin{aligned}
    \texttt{DLOT} = \inf_{\substack{\pi^s, \pi^f }} &\iint |\xi_{\text{orig}}(\mathbf{h}_{\text{orig}}, \mathbf{p}_{\text{orig}}) - \xi_{\text{pert}}(\mathbf{h}_{\text{pert}}^{(i)}, \mathbf{p}_{\text{pert}}^{(i)})|^p d\pi^s d\pi^f \\
    &+ \sum_{j=1}^2 \lambda_j \text{KL}\left(\pi^s_{\#j} \otimes \pi^f_{\#j} \big\| \mu^{s}_j \otimes \mu^{f}_{j}\right).
\end{aligned}
\nonumber
\end{equation}
To tolerate partial mass mismatch and enhance robustness, KL-divergence terms penalize deviations of the marginal distributions of $\pi^s$ and $\pi^f$ from their empirical counterparts $\mu^{s}$ and $\mu^{f}$.
A mass constraint $m(\pi^s) = m(\pi^f)$ is imposed to ensure transport consistency across levels.
We present the derivation details of solving \texttt{DLOT} problem in Appendix A.
The consistency reward is set as \texttt{DLOT} distance between as:
\begin{equation}
\small
    R_{\text{dist}}^{(i)} = -\texttt{DLOT}_{\lambda_1,\lambda_2}\left(\mathbb{X}_{\text{orig}}, \mathbb{X}_{\text{pert}}^{\prime(i)}\right),
\nonumber
\end{equation}
% Hyperparameters $\lambda_1$ and $\lambda_2$ modulate the trade-off between transport fidelity and marginal alignment strictness.
% The merit of adopting \texttt{DLOT} lies in its ability to relax marginal constraints via KL divergence, enhancing robustness to localized perturbations and aligning with the goal of population-level stealthiness.
% %
% Moreover, by jointly aligning inter-sequence samples and intra-sequence features, it enables multi-granular distributional matching, allowing perturbed sequences to preserve global behavioral patterns while mimicking fine-grained user intent transitions.

% 优势
% The merit of adopting \texttt{DLOT} lies in: i) By relaxing marginal constraints via KL divergence, it enhances robustness to localized perturbations and mitigates over penalization of minor anomalies, aligning with the goal of population-level stealthiness.
% %
% ii) Its joint alignment of inter-sequence samples and intra-sequence features enables multi-granular distributional matching, allowing perturbed sequences to retain global behavioral consistency while emulating fine-grained user

\subsection{Constrained Group Relative Learning}

To jointly optimize the multi-objective rewards under stealth-aware constraints, we introduce a two-stage optimization paradigm, which enables progressive learning of perturbation strategies by first distilling attack-effective behaviors and then aligning with distributional constraints.

\nosection{Constrained Reinforcement with Dynamic Barrier}
We model the perturbation decision process as a bi-level reinforcement learning problem, where the policy aims to maximize the inversion rewards while satisfying stealth constraints.
We formulate the bi-level objective for each mask step, where each reward or constraint term is defined as the expected discounted return:
\begin{equation}
\small
\begin{split}
&\max_{\theta} \mathcal{J}_{\text{dir}}(\psi) + \mathcal{J}_{\text{div}}(\psi) \quad \text{s.t.} \quad \mathcal{J}_{\text{dist}}(\psi) \leq \rho_{st} \\
\mathcal{J}_{r}(\psi) &= \mathbb{E}_{\tau \sim \pi_{\psi}} \left[ \sum_{t=0}^{T} \gamma^t R_r(s_t, a_t, s_{t+1}). \right], r \in \{\text{dir}, \text{div}, \text{dist}\}
\end{split}
\nonumber
\end{equation}
Here, $s_t$ denotes the state at step $t$ representing the partially perturbed sequence $\textbf{s}^{'(i)}$, $a_t$ represents the action of masking, and $\gamma^t$ is the discount factor.
The constraint $\mathcal{J}_{\text{dist}}(\psi) \leq \rho_{st}$ enforces stealthiness by bounding the expected distribution consistency reward.
It is worth noting that $\rho_{st}$ is a threshold dynamically derived from the distribution of trajectories in group relative policy, which ensures the stealth bound adapts to the evolving policy and group-wise sequence characteristics.
To solve this problem, we rewrite the problem as a min-max Lagrangian formulation:
\begin{equation}
\small
\mathcal{L}(\psi, \delta) = \mathcal{J}_{\text{dir}}(\psi) + \mathcal{J}_{\text{div}}(\psi) - \delta \cdot (\mathcal{J}_{\text{dist}}(\psi) - \rho_{st}), \quad \delta \geq 0
\nonumber
\end{equation}

We compute the policy gradients for each term via the policy gradient theorem, and further derive the policy gradient $\nabla_{\psi} \mathcal{L}$ for perturbation masker $\mathcal{M}_{\psi}$:
\begin{align}
\small
% \nabla_\psi J_{R_r}(\psi) &= \mathbb{E}_{\tau \sim \pi_\psi} \left[ \sum_t \nabla_\pi \log \pi_\psi(a_t|s_t) \cdot \hat{A}_{R_r}(s_t, a_t) \right], \quad r \in \{\text{dir}, \text{div}, \text{dist}\}\\
\nabla_{\psi} \mathcal{L} &= \nabla_{\psi} \mathcal{J}_{\text{dir}} + \nabla_{\psi} \mathcal{J}_{\text{div}} - \delta \cdot \nabla_{\psi} \mathcal{J}_{\text{dist}}.
\nonumber
\end{align}

Different from static constraints, we tend to dynamically adjust the penalty term, based on real-time constraint violation and gradient alignment to ensure a balanced optimization.
When gradients of $\mathcal{J}_{\text{dir}/\text{div}}$ and $\mathcal{J}_{\text{dist}}$ conflict, the numerator $\delta$ reduces to prioritize attack efficacy.
Severe stealth violations increase $\delta$ to suppress detectable perturbations.
Based on the dynamic barrier design \cite{gong2021bi}, we can give out the closed-form expression as
\begin{equation}
    \delta = \left[ \frac{ 
    \mathcal{J}_{\text{dist}} - \rho_{st} - \nabla_{\psi} \mathcal{J}_{\text{dist}}^\top \nabla_{\psi}(\mathcal{J}_{\text{dir}} + \mathcal{J}_{\text{div}}) 
    }{ 
        \left\| \nabla_{\psi} \mathcal{J}_{\text{dist}} \right\|^2 + \kappa 
    } \right]_+.
    \nonumber
\end{equation}
% \begin{equation}
% \delta = \left[
%   \frac{
%     \mathcal{J}_{R_{\text{dist}}}(\psi) - \rho_{st} - g_{\text{dist}}^\top g_{\text{rwd}}
%   }{
%     \|g_{\text{dist}}\|^2 + \kappa
%   }
% \right]_+,
% \nonumber
% \end{equation}
$\kappa > 0$ is for numerical stability. Then policy gradients are
\begin{equation}
\scriptsize
\begin{aligned}
\nabla_{\psi}(\mathcal{J}_{\text{dir}} + \mathcal{J}_{\text{div}}) &= \mathbb{E}_\tau \left[ \sum_t \nabla_\psi \log \pi_\psi(a_t|s_t) \cdot (\hat{A}_{R_{dir}} + \hat{A}_{R_{div}}) \right], \\
\nabla_{\psi} \mathcal{J}_{\text{dist}}&= \mathbb{E}_\tau \left[ \sum_t \nabla_\psi \log \pi_\psi(a_t|s_t) \cdot \hat{A}_{R_{dist}}(s_t, a_t) \right].
\end{aligned}
\nonumber
\end{equation}
% $
% \delta = \left[
%   \frac{
%     \mathcal{J}_{R_{dist}}(\psi) - \rho_{st} 
%     - \left(
%         \mathbb{E}_\tau \left[ \sum_t \nabla_\psi \log \pi_\psi(a_t|s_t) \cdot \hat{A}_C(s_t, a_t) \right]
%       \right)^\top
%       \left(
%         \mathbb{E}_\tau \left[ \sum_t \nabla_\psi \log \pi_\psi(a_t|s_t) \cdot (\hat{A}_{R_{dir}} + \hat{A}_{R_{div}}) \right]
%       \right)
%   }{
%     \left\| 
%       \mathbb{E}_\tau \left[ \sum_t \nabla_\psi \log \pi_\psi(a_t|s_t) \cdot \hat{A}_{\mathcal{J}_{R_{dist}}}(s_t, a_t) 
%     \right] \right\|^2 + \kappa
%   }
% \right]_+
% % \nonumber
% $
%
This ensures the policy prioritizes stealthiness only when constraints are violated, balancing efficacy and detectability without sacrificing convergence stability.
The perturbation policy is updated using gradient ascent on the Lagrangian as 
$
\psi_{t+1} = \psi_t + \eta \left( \nabla_{\psi} \mathcal{J}_{\text{dir}} + \nabla_{\psi} \mathcal{J}_{\text{div}} - \delta \cdot \nabla_{\psi} \mathcal{J}_{\text{dist}} \right)
$, where $\eta$ denotes the learning rate.
Detailed derivations of this part are in Appendix B.
% This dynamic barrier ensures the constraint is softly enforced throughout training, and the reward gradients are adaptively corrected according to the degree of stealth violation.
%

\nosection{Group Relative Optimization}
To stabilize optimization and accelerate convergence, inspired by the GRPO paradigm \cite{liu2024deepseek}, we further reform the constrained reinforcement learning with a group relative strategy.
Formally, we divide the training into two stages, i.e., \textit{localization stage} and \textit{constrained fine-tuning stage}.
In the localization stage, we train the perturbation masker using only the pattern-level rewards $R_{\text{dir}}$ and $R_{\text{div}}$.
This stage is analogous to supervised fine-tuning, allowing the policy to explore effective inversion behaviors in an unconstrained space.
We obtain the masker policy with the pure attacking goal as $\pi_{\text{att}}$.
In the constrained stage, we introduce the distribution consistency reward $R_{\text{dist}}$ as a constraint and perform constrained reinforcement learning with with dynamic barrier, guided by a GRPO-based surrogate objective.
In each masking step, we sample a group of $G$ trajectories $\{o_{i,t}, r_{i,t}\}_{i=1}^G$ under the current policy $\pi_{\psi}$, where $o_{i,t}$ is the trajectory and $r_{i,t}$ is the reward aggregated from multi-objective signals.
We obtain the group-wise baseline $\mu_{G,t} = \frac{1}{G}\sum_{i=1}^G r_{i,t}$ and reward standard deviation $\sigma_{G,t} = \sqrt{ \frac{1}{G-1} \sum_{i=1}^{G} (r_{i,t} -\mu_{G,t})^2 }$. 
% \begin{equation}
% \small
%     \mu_G = \frac{1}{G} \sum_{i=1}^{G} r_i, \quad \sigma_G = \sqrt{ \frac{1}{G-1} \sum_{i=1}^{G} (r_i - \mu_G)^2 }.
% \nonumber
% \end{equation}
%
Besides, we can also set the dynamic stealthy threshold for the constraint reward as $\rho_{st} = \mu_{G}^{dist} + \lambda_{st} \cdot \sigma_{G}^{dist}$.
We construct the normalized advantage estimate $
\hat{A}_{i,t} = \frac{r_{i,t} - \mu_{G,t}}{\sigma_{G,t} + \epsilon}
$, where $\epsilon$ is a positive constant for numerical stability.
Finally, the policy is optimized with a clipped surrogate objective:
\begin{equation}
% \small
% \scriptsize
% \scriptstyle
\begin{aligned}
\mathcal{J}(\psi) &= \mathbb{E}_\tau \left[\frac{1}{G} \sum_{i=1}^{G} \frac{1}{\|o_i\|} \sum_{t=1}^{\|o_i\|} \min \left( \frac{\pi_{\psi}(o_{i,t})}{\pi_{\text{att}}(o_{i,t})} \hat{A}_{i,t}, \right. \right. \\
&\quad \left. \left. \text{clip} \left( \frac{\pi_{\psi}(o_{i,t})}{\pi_{\text{att}}(o_{i,t})}, 1 - \epsilon, 1 + \epsilon \right) \hat{A}_{i,t} \right) \right]
\end{aligned}
\nonumber
\end{equation}
This group relative normalization enhances training stability by reducing sensitivity to outliers and scales advantage estimates adaptively across trajectories.
Combining this with bi-level optimization, C-GRRL supports efficient and stable learning of stealthy, high-impact perturbation strategies.

\section{Empirical Study}

\subsection{Experimental Setups}

\nosection{Datasets}
Following \cite{defending}, we evaluate on three datasets, i.e. ML-1M, ML-20M, and Amazon Beauty, and preprocess them according to \cite{bert4reccc}. The last two items in each sequence are reserved for validation and testing, with the rest assigned to training. The details of datasets are shown in Appendix D.

% \begin{table}
% \small
% \renewcommand\arraystretch{0.6}
% \centering
% \caption{Statistics on Datasets.}
% \label{tab:dataset}
% \begin{tabular}{cccccc}
%     \toprule
%     \textbf{Datasets} & \textbf{\#User} & \textbf{\#Item} & \textbf{Avg.len} & \textbf{Sparsity} \\
%     \midrule
%     ML-1M  & 6,040 & 3,416 & 165 & 95.2\%\\
%     \midrule
%     Beauty & 22,363  & 12,101 & 9 & 99.9\%\\
%     \midrule
%     ML-20M & 138,493 & 18,345 & 144 & 99.2\%\\
%     \bottomrule
% \end{tabular}
% \vspace{-0.1 in}
% \end{table}

\nosection{Backbone Recommender Systems}
We choose two representative SR: 
1) \textbf{NARM} \cite{li2017neural} integrates a GRU-based global and local encoder for sequential modeling. 
2) \textbf{BERT4Rec} \cite{bert4reccc} employs a transformer-based architecture with bidirectional self-attention. 

\nosection{Comparison Methods}
For fair comparison, we set all attack SOTAs under the same setting, \ie PPA against white-box SR: 1) \textbf{Popular Attack} injects sequences with target items and filler popular items.
2) \textbf{Random Attack} is similar to the popular attack, but randomly selects filler items.
3) \textbf{SimAlter Attack} \cite{blackbox} constructs adversarial sequences with semantic associations by calculating the gradient perturbation direction. 
4) \textbf{Replace Attack} \cite{defending} identifies fragile items by the sequence gradient information and replaces them with adversarial candidates.
5) \textbf{SSL Attack} \cite{wang2023poisoning} realizes data poisoning against SR with generative adversarial networks, and we adapt it into our white-box setting for fair comparison.

\nosection{Evaluation Protocols}
We evaluate with the exposure ratio of the target item, which is measured with Hit Ratio (HR), Normalized Discounted Cumulative Gain (NDCG), and Mean Reciprocal Rank (MRR).
We set the cut-off of the ranked list as 1, 5, and 10.
For all the experiments, we repeat them five times and report the average results. The statistical significance tests are conducted by performing $t$-tests.

% 一共5个baseline
% RandomAlter
% PopularAlter
% SimAlter
% Replace
% Influence

\nosection{Implementation Details}
We present the implementation details and parameter settings in Appendix D.

\subsection{Experimental Results and Analysis}

\nosection{Overall Attack Performance (RQ1)}
We evaluate the attack performance of \modelname~and baselines on NARM and Bert4Rec for three datasets.
From Table \ref{tab:perform}, we find: 1) \modelname~ significantly enhances target item exposure across diverse scenarios, delivering nearly ten times higher exposure compared to \textit{Pure}.
Specifically, it outperforms the best baselines by over $20\%$ across all metrics. 
2) Dense datasets favor appending strategies (\eg Random, Popular), whereas short sequences exhibit heightened susceptibility to substitution strategies (\eg SimAlter, Replace).
The divergence emerges as longer sequences intensify positional bias in attention mechanisms, while shorter sequences' dependence on sparse high-impact features elevates substitution risks.
3) On the most challenging dataset, \ie \textit{Beauty}, all baselines exhibit poor attack performance, \eg the well-performed SOTA \textit{SSL Attack} only achieve 0.0109 on HR@10.
In contrast, \modelname~achieved 0.2601 on \textit{HR@10} and 0.1038 on \textit{NDCG@10}, highlighting the superior efficacy of deep pattern extraction over gradient-based strategies.
% 3) The baselines perform poorly on \textit{Beauty}. For Bert4Rec, baseline methods showed near-zero performance metrics with second-best results achieving 0.0048 on \textit{HR@10} and 0.0017 on \textit{NDCG@10}. Our method maintained 0.2601 on \textit{HR@10} and 0.1038 on \textit{NDCG@10}, which indicates that deep pattern extraction outperforms simple gradient-based strategies.

\begin{table*}[t]
% \small
% \renewcommand\arraystretch{0.6}
\centering
% \vspace{-0.1in}
\resizebox{1.0\linewidth}{!}{
\begin{tabular}{c|cccccc|cccccc}
\toprule
 \multirow{2}{*}{\textbf{Attack}}
& \multicolumn{6}{c|}{\textbf{NARM}}
& \multicolumn{6}{c}{\textbf{Bert4Rec}}\\
 % \cmidrule(lr){1-1} \cmidrule(lr){2-4} \cmidrule(lr){5-7}
\cmidrule(lr){2-7} \cmidrule(lr){8-13}

% \multirow{2}{*}{ER@10 (AZ, IJCAI)} & \multicolumn{2}{c}{$\mu=0.1\%$} & \multicolumn{2}{c}{$\mu=1.0\%$} & \multicolumn{2}{c}{$\mu=10\%$} & \multicolumn{2}{c}{$\mu=0.1\%$} & \multicolumn{2}{c}{$\mu=1.0\%$} & \multicolumn{2}{c}{$\mu=10\%$} \\
& HR@1 & HR@5 & HR@10 & NDCG@5 & NDCG@10 & MRR & HR@1 & HR@5 & HR@10 & NDCG@5 & NDCG@10 & MRR \\
 \midrule
\multicolumn{13}{c}{\textbf{ML-1M}} \\
\midrule
Pure & 0.0038 & 0.0109 & 0.0174 & 0.0075 & 0.0096 & 0.0098 & 0.0000 & 0.0076 & 0.0128 & 0.0034 & 0.0051 & 0.0062\\
Popular & 0.0128 & 0.0428 & 0.0695 & 0.0280 & 0.0365 & 0.0406 & 0.0105 & \underline{0.0632} & \underline{0.1065} & \underline{0.0372} & \underline{0.0515} & 0.0452\\
Random & 0.0104 & 0.0326 & 0.0560 & 0.0217 & 0.0291 & 0.0333 & 0.0043 & 0.0305 & 0.0560 & 0.0173 & 0.0255 & 0.0254\\
SimAlter & 0.0143 & 0.0459 & 0.0658 & 0.0306 & 0.0369 & 0.0360 & 0.0102 & 0.0385 & 0.0563 & 0.0271 & 0.0328 & 0.0324\\
Replace & \underline{0.0164} & \underline{0.0575} & 0.1027 & \underline{0.0369} & \underline{0.0515} & 0.0514 & \underline{0.0155} & 0.0569 & 0.0938 & 0.0365 & 0.0493 & \underline{0.0512}\\
SSLAttack & 0.0148 & 0.0565 & \underline{0.1069} & 0.0350 & 0.0510 & \underline{0.0564} & 0.0145 & 0.0578 & 0.0970 & 0.0356 & 0.0483 & 0.0453\\
\rowcolor{gray!18} \modelname & \textbf{0.0305} & \textbf{0.0855} & \textbf{0.1428} & \textbf{0.0580} & \textbf{0.0765} & \textbf{0.0749} & \textbf{0.0187} & \textbf{0.0811} & \textbf{0.1492} & \textbf{0.0448} & \textbf{0.0666} & \textbf{0.0593}\\
%Impv. & 46.23\% & 48.70\% & 39.05\% & 57.18\% & 48.54\% & 45.72\% & 20.65\% & 28.32\% & 40.09\% & 20.43\% & 29.32\% & 31.19\%\\
\midrule

\multicolumn{13}{c}{\textbf{Beauty}} \\
\midrule

Pure & 0.0003 & 0.0015 & 0.0033 & 0.0008 & 0.0014 & 0.0021 & 0.0000 & 0.0000 & 0.0000 & 0.0000 & 0.0000 & 0.0022\\
Popular & 0.0005 & 0.0018 & 0.0065 & 0.0009 & 0.0023 & 0.0032 & 0.0000 & 0.0000 & 0.0000 & 0.0000 & 0.0000 & 0.0026\\
Random & 0.0010 & 0.0051 & 0.0103 & 0.0029 & 0.0042 & 0.0049 & 0.0000 & 0.0000 & 0.0001 & 0.0000 & 0.0000 & 0.0031\\
SimAlter & 0.0019 & 0.0093 & 0.0178 & 0.0055 & 0.0082 & 0.0116 & 0.0000 & 0.0013 & 0.0048 & 0.0006 & 0.0017 & 0.0068\\
Replace & 0.0027 & 0.0151 & 0.0608 & 0.0070 & 0.0216 & 0.0228 & 0.0000 & 0.0000 & 0.0000 & 0.0000 & 0.0000 & 0.0034\\
SSLAttack & \underline{0.0095} & \underline{0.0354} & \underline{0.0751} & \underline{0.0226} & \underline{0.0296} & \underline{0.0270} & \underline{0.0022} & \underline{0.0074} & \underline{0.0109} & \underline{0.0049} & \underline{0.0060} & \underline{0.0097}\\
\rowcolor{gray!18} \modelname & \textbf{0.0424} & \textbf{0.1080} & \textbf{0.1504} & \textbf{0.0759} & \textbf{0.0895} & \textbf{0.0801} & \textbf{0.0076} & \textbf{0.1191} & \textbf{0.2601} & \textbf{0.0582} & \textbf{0.1038} & \textbf{0.0773} \\
\midrule

\multicolumn{13}{c}{\textbf{ML-20M}} \\
\midrule

Pure & 0.0020 & 0.0108 & 0.0215 & 0.0063 & 0.0097 & 0.0136 & 0.0001 & 0.0015 & 0.0046 & 0.0007 & 0.0018 & 0.0070\\
Popular & \underline{0.0516} & \underline{0.1516} & \underline{0.2207} & \underline{0.1024} & \underline{0.1247} & \underline{0.1074} & 0.0201 & \underline{0.1542} & \underline{0.2431} & \underline{0.0905} & \underline{0.1171} & \underline{0.0945}\\
Random & 0.0411 & 0.1287 & 0.2080 & 0.0851 & 0.1096 & 0.1019 & 0.0193 & 0.1062 & 0.1634 & 0.0746 & 0.0930 & 0.0892\\
SimAlter & 0.0318 & 0.1204 & 0.2048 & 0.0756 & 0.1036 & 0.0955 & \underline{0.0235} & 0.0815 & 0.1398 & 0.0523 & 0.0710 & 0.0662\\
Replace & 0.0356 & 0.1243 & 0.1906 & 0.0799 & 0.1012 & 0.0921 & 0.0084 & 0.0455 & 0.0872 & 0.0269 & 0.0402 & 0.0420\\
SSLAttack & 0.0437 & 0.1202 & 0.1743 & 0.0825 & 0.0999 & 0.0893 & 0.0217 & 0.0723 & 0.1135 & 0.0473 & 0.0605 & 0.0583\\
\rowcolor{gray!18} \modelname & \textbf{0.0585} & \textbf{0.1905} & \textbf{0.3041} & \textbf{0.1243} & \textbf{0.1608} & \textbf{0.1392} & \textbf{0.0241} & \textbf{0.1621} & \textbf{0.2628} & \textbf{0.0932} & \textbf{0.1255} & \textbf{0.1025}\\
\bottomrule

\end{tabular}
}
\vspace{-0.1 in}
\caption{The overall performance on three datasets. The best results are boldfaced, and the second-best results are underlined. All improvements are significant with $p$-value $<$ 0.05 based on $t$-tests.}
\label{tab:perform}
\vspace{-0.1 in}
\end{table*}

\begin{table}[t]
\renewcommand\arraystretch{0.9}
\centering
\vspace{-0.1in}
\resizebox{1.0\linewidth}{!}{
\begin{tabular}{c|cccc|cccc}
\toprule
 \multirow{3}{*}{\textbf{Attack}}
& \multicolumn{4}{c|}{\textbf{NARM}}
& \multicolumn{4}{c}{\textbf{Bert4Rec}}\\
 % \cmidrule(lr){1-1} \cmidrule(lr){2-4} \cmidrule(lr){5-7}
\cmidrule(lr){2-5} \cmidrule(lr){6-9}

% \multirow{2}{*}{ER@10 (AZ, IJCAI)} & \multicolumn{2}{c}{$\mu=0.1\%$} & \multicolumn{2}{c}{$\mu=1.0\%$} & \multicolumn{2}{c}{$\mu=10\%$} & \multicolumn{2}{c}{$\mu=0.1\%$} & \multicolumn{2}{c}{$\mu=1.0\%$} & \multicolumn{2}{c}{$\mu=10\%$} \\
& \multicolumn{2}{c|}{ML-1M} & \multicolumn{2}{c|}{Beauty} & \multicolumn{2}{c|}{ML-1M} & \multicolumn{2}{c}{Beauty}\\
\cmidrule(lr){2-3} \cmidrule(lr){4-5}\cmidrule(lr){6-7} \cmidrule(lr){8-9} 
& H@10 & N@10 & H@10 & N@10 & H@10 & N@10 & H@10 & N@10  \\

 \midrule

w/o dir & 0.0117 & 0.0024 & 0.5105 & 0.3663 & 0.0140 & 0.0056 & 0.6307 & 0.4812\\
w/o div & 0.0143 & 0.0055 & 0.5547 & 0.4318 & 0.0151 & 0.0063 & 0.6603 & 0.5941\\
w/o pv & 0.0025 & 0.0013 & 0.5051 & 0.3538 & 0.0053 & 0.0019 & 0.5044 & 0.2859\\
w/o dist & 0.0271 & 0.0116 & 0.6291 & 0.5221 & 0.0212 & 0.0083 & 0.7461 & 0.7034\\
\rowcolor{gray!18} \modelname &  \textbf{0.0157} & \textbf{0.0063} & \textbf{0.5633} & \textbf{0.4439} & \textbf{0.0186} & \textbf{0.0068} & \textbf{0.6885} & \textbf{0.6285}\\

\bottomrule

\end{tabular}
}
\vspace{-0.05 in}
\caption{Ablation studies on each reward.}
\label{tab:ablation}
\vspace{-0.15 in}
\end{table}

\nosection{Ablation Studies (RQ2)}
To evaluate impact of each reward in policy, we design variants as:
(a) \textit{w/o dir} removes the directionality reward.
(b) \textit{w/o div} removes the diversity reward.
(c) \textit{w/o pv} removes the whole pattern inversion reward.
(d) \textit{w/o dist} removes the distribution consistency reward.
From Table \ref{tab:ablation}, we conclude: 1) \textit{w/o dir} and \textit{w/o div} both outperform \textit{w/o pv}, showing that either the directionality or diversity reward contributes to the promotion.
2) \modelname~performs better than \textit{w/o dir} and \textit{w/o div}, showing that relying solely on either pattern reward has its limitations.
Utilizing only $R_{dir}$ leads to the clustering of inverted patterns in the representation space, while solely on $R_{div}$ results in insufficient disturbance of the sub-patterns.
3) The constrained reward $R_{dist}$ decreases attack effectiveness on both datasets, but its impact varies.
This is due to different behavioral characteristics, \ie high randomness in user behavior makes patterns more fragile, while stable preferences allow effective feature injection even under covert conditions.

%% Part 1: 3 个 reward 分别消融:
%% (1) -inversion (2) -diversity (3) -inv+div (4) -ot

\begin{figure}[t]
\centering
\vspace{-0.1in}
\subfigure[Bert4Rec]{\label{fig:subfig:re-1}
\resizebox{1.0\hsize}{!}
{\includegraphics[width=0.49\linewidth]{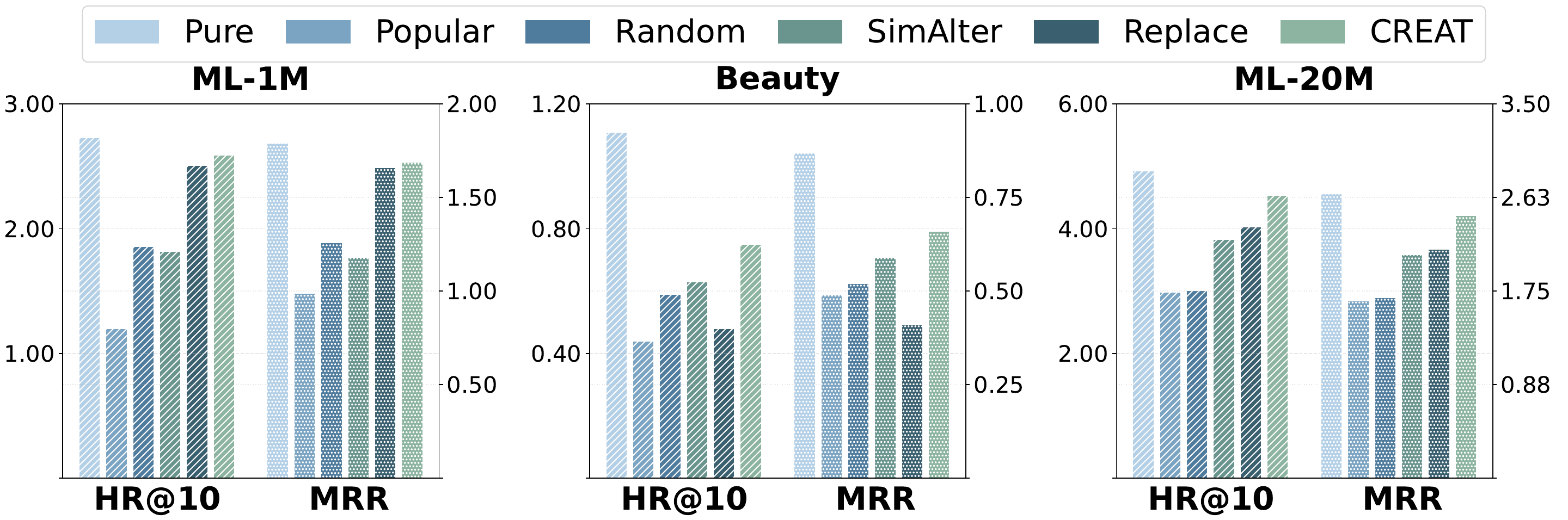}}}
% \hspace{0.01\linewidth}
\subfigure[NARM]{\label{fig:subfig:re-2}
\resizebox{1.0\hsize}{!}
{\includegraphics[width=0.49\linewidth]{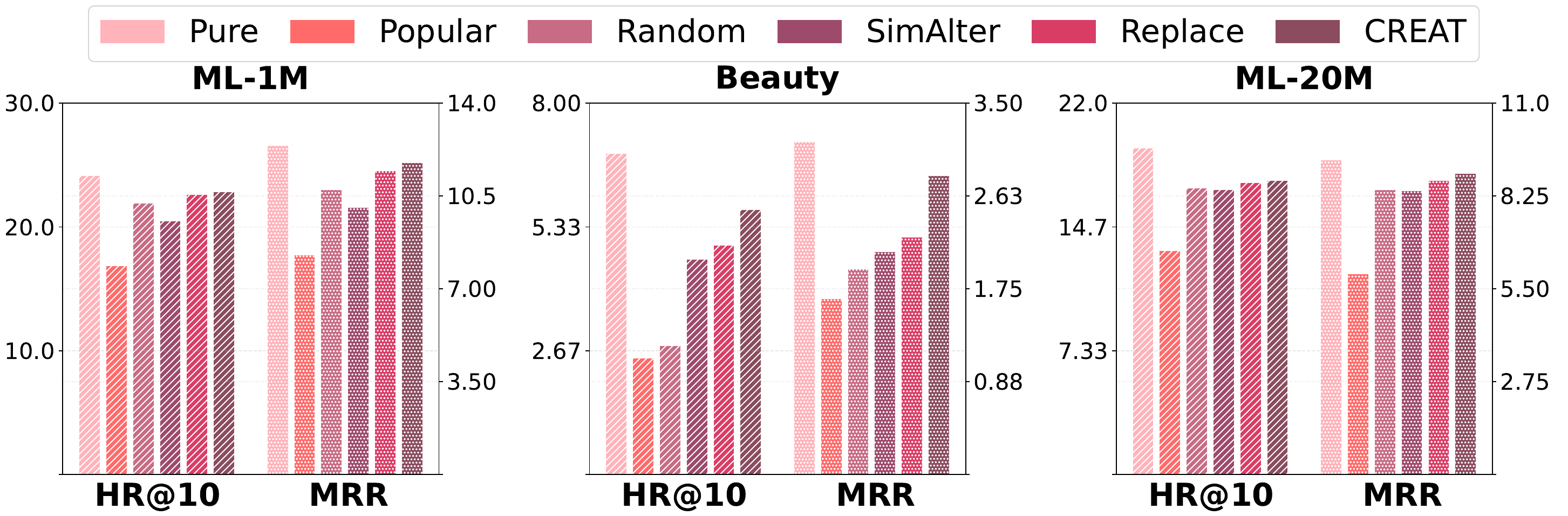}}}
\vspace{-0.2 in}
\caption{Effects on recommendation accuracy (\%).}
\label{fig:rec_perform}
\vspace{-0.25 in}
\end{figure}

\begin{figure*}[t]
\centering
\includegraphics[width=1.0\linewidth]{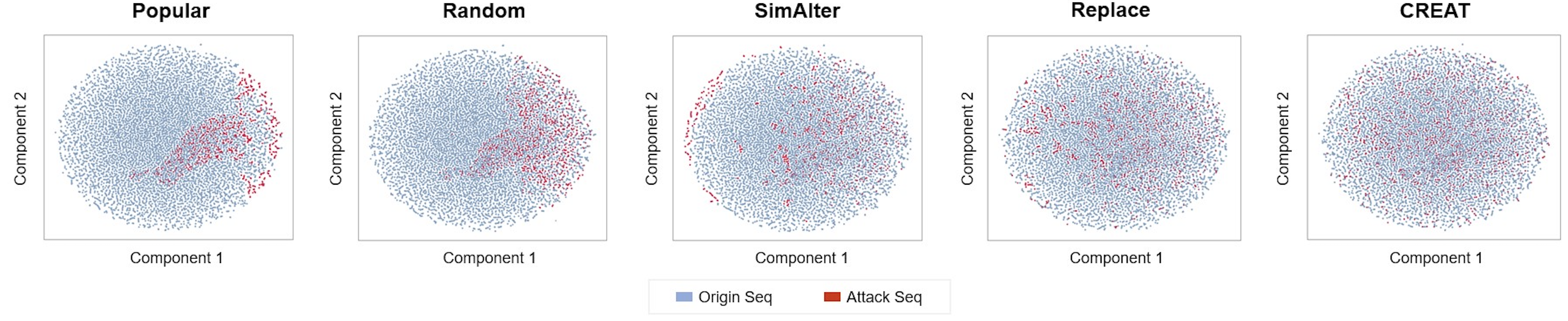}
\vspace{-0.25 in}
\caption{The t-SNE visualization of original sequences and polluted sequences.} 
\label{fig:tsne}
\vspace{-0.2 in}
\end{figure*}

\nosection{Stealth Verification (RQ3)}
%%【TSNE】
We evaluate the stealthiness of \modelname~from three aspects.
\textit{First}, we present the performance of each attack under the defense SOTA methods in Appendix E, where we prove \modelname~still outperforms other attacks even under the strongest defenses.
\textit{Second}, we evaluate the side effects that each PPA brings to the overall SR performance.
We present the recommendation accuracy of all users on both backbones before and after PPA in Figure \ref{fig:rec_perform}.
From it, we observe all attacks damage the performance of SR.
Compared to SOTAs, \modelname~results in slighter accuracy degradation across all datasets, indicating that: 1) \modelname~efficiently attacks by uncovering finer-grained sequential pattern correlations with minimal disturbance. 2) Since \modelname~causes less degradation in the overall accuracy, it is less likely to trigger detection mechanisms that rely on fluctuations in SR performance, thereby making the attack more stealthy.
\textit{Third}, we employ t-SNE \cite{van2008visualizing} to visualize the latent distribution of original sequences and polluted sequences.
From Figure \ref{fig:tsne}, we find:
1) the adversarial samples from \textit{Popular} and \textit{Random} are distinguishable from original samples, while \textit{SimAlter} with feature alignment improves stealth but still has detectable anomalies.
2) Compared to the best SOTA \textit{Replace}, \modelname~shows further enhancement in stealth, achieving indistinguishability between polluted and clean sequences, indicating high integration in spatial density, local clustering patterns, and edge distribution features.

\begin{figure}[t]
\centering
\includegraphics[width=0.95\linewidth]{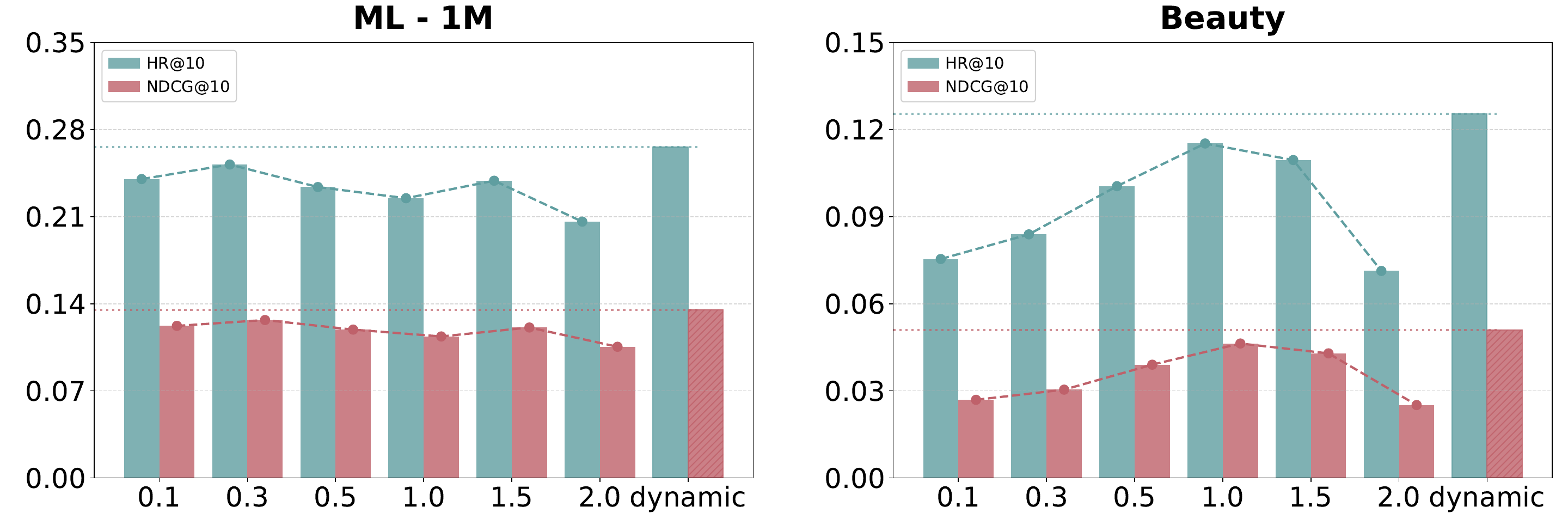}
\vspace{-0.12 in}
\caption{The effects of the dynamic barrier.} 
\label{fig:dynamic}
\vspace{-0.18 in}
\end{figure}

\nosection{Adaptability of Dynamic Barrier (RQ4)}
%%【折线图/柱状图】
%% 对bi-level的\lambda进行研究，设置若干个（3个）定值\lambda和我们的模型进行 attack performance的比较
To justify necessity of dynamically adjusting constraint penalty, we replace the constrained reinforcement with \textit{dynamic barrier} by fixed penalty coefficients.
From Figure \ref{fig:dynamic}, we observe that setting the penalty to static values, \ie ranging from 0.1 to 2.0, leads to suboptimal or unstable performance.
Contrastingly, \modelname~achieves the best performance, indicating the ability of dynamic barrier to adaptively adjust the penalty in response to constraint violations and optimization dynamics.

% \begin{wrapfigure}{R}{0.48\textwidth}
% \vspace{-20pt}
% \begin{center}
% \includegraphics[width=0.48\textwidth]{pic/dynamic.png}
% \end{center}
% \vspace{-5pt}
% \caption{The effects of dynamic barrier.}
% \label{fig:dynamic}
% \vspace{-5pt}
% \end{wrapfigure}

\begin{figure}[t]
\centering
\includegraphics[width=1.0\linewidth]{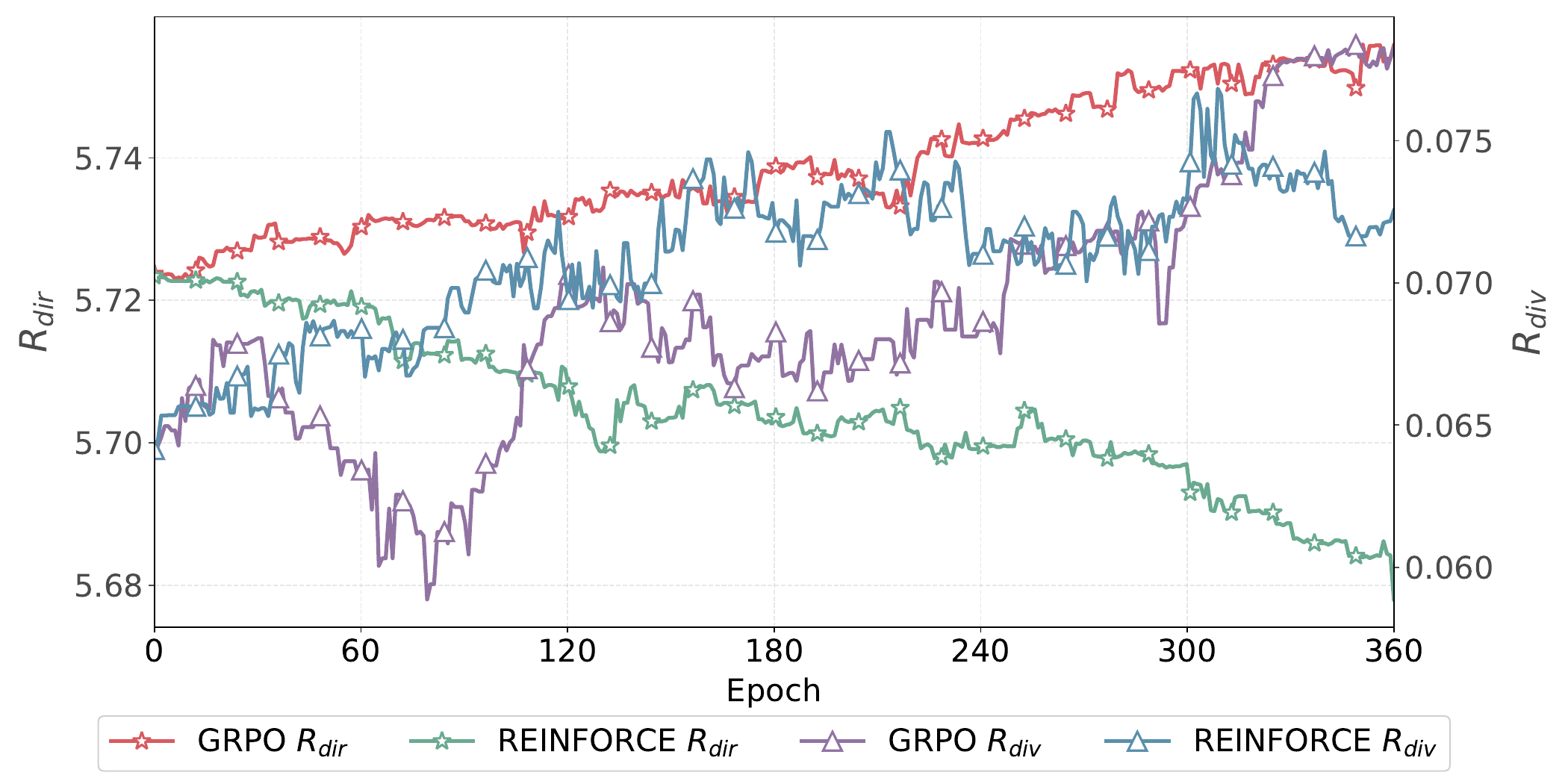}
\vspace{-0.2 in}
\caption{The convergence of \modelname.} 
\label{fig:convergence}
\vspace{-0.15 in}
\end{figure}

\nosection{GRPO Convergence (RQ5)}
We prove GRPO's convergence through first-360-epoch reward trajectories, compared with traditional REINFORCE.
Figure \ref{fig:convergence} shows REINFORCE preserves consistency reward but suffers persistent pattern inversion reward decline due to its lack of group-wise baseline normalization for reward functions, whereas GRPO achieves sustained pattern inversion reward growth stabilizing post-epoch-300.
Despite initial volatility, GRPO's consistency reward stabilizes post-epoch-120 and converges post-epoch-330, confirming its capability.
%% GRPO消融对于收敛性的影响 【折线图】
%% (1) tradition_rl (2) grpo [体现Convergence，不用体现attack performance]

\nosection{Effect of Target Item Popularity (RQ6)} 
%% 单独弄一张表 head medium tail
%% 这里的baseline只放：black-box defend ours
To investigate the effect of target item popularity on attacks, we conduct more experiments.
From Table \ref{tab:popularity}, we find:
1) \textit{SimAlter} and \textit{Replace} exhibit significantly weaker tail-item attack capabilities than \modelname, demonstrating near-zero efficacy across all datasets.
% The degraded perturbation effectiveness stems from the distortion of the gradient signal induced by sparse interaction data and the displacement of the embedding space, which hinder adversarial perturbation generation and item substitution mechanisms.
2) \modelname~exhibits superior exposures across all range of popularity.
Our attack demonstrates a greater relative improvement on lower-popularity targets, with tail items exhibit higher enhancement than medium and head targets.
Since attackers in real scenarios typically focus on increasing the illicit exposure of long-tail items rather than already popular ones, this advantage has practical significance.

\begin{table}[t]
\footnotesize
\centering
% \vspace{-0.1in}
\resizebox{1.0\linewidth}{!}{
\begin{tabular}{c|ccc|ccc}
\toprule
 \multirow{2}{*}{\textbf{Attack}}
& \multicolumn{3}{c|}{\textbf{NARM}}
& \multicolumn{3}{c}{\textbf{Bert4Rec}}\\
 % \cmidrule(lr){1-1} \cmidrule(lr){2-4} \cmidrule(lr){5-7}
\cmidrule(lr){2-4} \cmidrule(lr){5-7}

% \multirow{2}{*}{ER@10 (AZ, IJCAI)} & \multicolumn{2}{c}{$\mu=0.1\%$} & \multicolumn{2}{c}{$\mu=1.0\%$} & \multicolumn{2}{c}{$\mu=10\%$} & \multicolumn{2}{c}{$\mu=0.1\%$} & \multicolumn{2}{c}{$\mu=1.0\%$} & \multicolumn{2}{c}{$\mu=10\%$} \\
& \multicolumn{1}{c|}{Head} & \multicolumn{1}{c|}{Medium} & \multicolumn{1}{c|}{Tail} & \multicolumn{1}{c|}{Head} & \multicolumn{1}{c|}{Medium} & \multicolumn{1}{c|}{Tail}\\
% \cmidrule(lr){2-3} \cmidrule(lr){4-5}\cmidrule(lr){6-7} \cmidrule(lr){8-9} \cmidrule(lr){10-11} \cmidrule(lr){12-13}
% & HR@10 & NDCG@10 & HR@10 & NDCG@10 & HR@10 & NDCG@10 & HR@10 & NDCG@10 & HR@10 & NDCG@10 & HR@10 & NDCG@10\\

 \midrule

\multicolumn{7}{c}{\textbf{ML-1M}}\\
\midrule

Pure &  0.0513 & 0.0174& 0.0000  & 0.2149  & 0.0128 & 0.0000 \\
SimAlter & 0.3079  & 0.0658  & 0.0000  & 0.3081  & 0.0563 & 0.0000 \\
Replace & \underline{0.3535}  & \underline{0.1027} & \underline{0.0036}  & \underline{0.4829}  & \underline{0.0938}  & 0.0000 \\
\rowcolor{gray!18} \modelname & \textbf{0.5664}  & \textbf{0.1428}  & \textbf{0.0451}  & \textbf{0.5228} & \textbf{0.1492} & \textbf{0.0707} \\
\midrule

\multicolumn{7}{c}{\textbf{Beauty}} \\
\midrule

Pure & 0.0472 & 0.0033 & 0.0000  & 0.0254 & 0.0000  & 0.0000 \\
SimAlter & 0.1504 & 0.0178 & \underline{0.0004} & \underline{0.5277} & \underline{0.0048}  & 0.0000 \\
Replace & \underline{0.3350}  & \underline{0.0608}  & 0.0000  & 0.4947 & 0.0000 & 0.0000 \\
\rowcolor{gray!18} \modelname & \textbf{0.4714}  & \textbf{0.1504}& \textbf{0.0102} & \textbf{0.6587}  & \textbf{0.2601}  & \textbf{0.1999} \\
\midrule

\multicolumn{7}{c}{\textbf{ML-20M}} \\
\midrule

Pure & 0.0974 & 0.0215 & 0.0000 & 0.2139  & 0.0046  & 0.0000 \\
SimAlter & \underline{0.4031}  & \underline{0.2048} & 0.0000 & 0.3401 & \underline{0.1398}  & 0.0000 \\
Replace & 0.3865  & 0.1906  & 0.0000  & \underline{0.3816}  & 0.0872  & 0.0000 \\
\rowcolor{gray!18} \modelname & \textbf{0.4317} & \textbf{0.3041}& \textbf{0.0953} & \textbf{0.4590} & \textbf{0.2628} & \textbf{0.1502} \\
\bottomrule

\end{tabular}
}
\vspace{-0.1 in}
\caption{Attack across various target popularity (HR@10).}
\label{tab:popularity}
\vspace{-0.15 in}
\end{table}

% \nosection{Parameter Analysis}
%% 这块随机应变，不需要在正文呈现所有的dataset和backbone
%% target item 替换的比例 （对attack的影响）【表格】

%% ot distance 阈值参数 （对stealth的影响）【TSNE】【KDE】

\section{Conclusion and Future Work}\label{conclusion}
We rethink the PPA against SR into a bi-level optimization problem with upper-level objective as attack intensity and lower-level as attack stealthiness.
In the proposed\modelname, we first design a pattern balanced rewarding policy to reverse key patterns and reduce observable deviations.
Then we implement a constrained group relative reinforcement learning, leveraging dynamic constraints and group-shared experience replay.
Extensive experiments demonstrate the effectiveness and stealth of \modelname.
\modelname~exhibit superior attack effectiveness, but there are currently no effective defenses against such attacks, which pose potential security risks to recommenders.
In the future, we will extend adaptive defenses against such attacks for recommender safety.

\section{Acknowledgments}
This work is supported by the National Natural Science Foundation of China (Grant No.72192823 and No.624B2131) and the Fundamental Research Funds for the Central Universities.
Xiaobo Xia is partially supported by the MoE Key Laboratory of Brain-inspired Intelligent Perception and Cognition at the University of Science and Technology of China (Grant No. 2421002).

% \bigskip

\bibliography{aaai2026}

\clearpage

\section*{Appendix}

\section{A \  Distribution Consistency Reward}
% \label{appendix:dlot}

This section presents the complete formulation and optimization procedure for the Dual-Level co-Optimal Transport (DLOT) reward, which we use to quantify semantic consistency between the original and perturbed sequences.
% The derivation is based on the \textit{Unbalanced CO-Optimal Transport (UCOOT)} framework~\cite{tran2023ucoot}.

\subsection{DLOT Formulation}

Let $\mathbf{s}$ denote the original user sequence and $\mathbf{s}^{'(i)}$ its perturbed counterpart. Their corresponding global sequence embeddings from the recommender encoder are:
\begin{equation}
\mathbf{h}_{\text{orig}} = \varphi_{\text{rec}}(\mathbf{s}), \quad \mathbf{h}_{\text{pert}}^{(i)} = \varphi_{\text{rec}}(\mathbf{s}^{'(i)}).
\end{equation}
In parallel, we extract local $k$-gram representations from each sequence:
\begin{equation}
\begin{aligned}
\mathbf{p}_{\text{orig}} = \left\{ \varphi_{\text{rec}}(\mathbf{s}[t:t+k]) \right\}_{t=1}^{L-k}, \\
\mathbf{p}_{\text{pert}}^{(i)} = \left\{ \varphi_{\text{rec}}(\mathbf{s}^{'(i)}[t:t+k]) \right\}_{t=1}^{L-k}.
\end{aligned}
\end{equation}

We define the sample-feature spaces as $\mathbb{X}_{\text{orig}} = (\mathbf{h}_{\text{orig}}, \mathbf{p}_{\text{orig}}, \xi_{\text{orig}})$ and $\mathbb{X}_{\text{pert}}^{(i)} = (\mathbf{h}_{\text{pert}}^{(i)}, \mathbf{p}_{\text{pert}}^{(i)}, \xi_{\text{pert}})$, where $\xi$ is a scalar interaction function, e.g., dot-product or neural similarity). Following the UCOOT formulation \cite{tran2023unbalanced}, the DLOT distance is defined as:
\begin{equation}
\begin{aligned}
\text{DLOT}_\lambda(\mathbb{X}_{\text{orig}}, \mathbb{X}_{\text{pert}}^{(i)}) = \min_{\pi^s, \pi^f} \Bigg[
\iint \left| \xi_{\text{orig}}(h_i, p_k) - \xi_{\text{pert}}(h_j, p_l) \right|^p  \\ d\pi^s(i,j) d\pi^f(k,l) \notag
 + \sum_{j=1}^2 \lambda_j \text{KL}(\pi^s_{\#j} \otimes \pi^f_{\#j} \,\|\, \mu_j^s \otimes \mu_j^f)
\Bigg],
\end{aligned}
\end{equation}
which subjects to the constraint $m(\pi^s) = m(\pi^f)$. This objective captures both global and local alignment under unbalanced settings, where marginal relaxation enables robustness to noise or perturbations.

The corresponding reward is defined as the negative DLOT value:
\begin{equation}
R_{\text{dist}}^{(i)} = - \text{DLOT}_\lambda(\mathbb{X}_{\text{orig}}, \mathbb{X}_{\text{pert}}^{(i)}),
\end{equation}
which encourages perturbed sequences to remain distributionally consistent with the original in both global and compositional representations.

\subsection{Optimization via BCD and Sinkhorn Algorithm}

To solve the DLOT objective efficiently, we adopt the \textit{Block Coordinate Descent (BCD)} algorithm proposed for \cite{tran2023unbalanced}, with alternating updates for the sample-level transport plan $\pi^s$ and feature-level plan $\pi^f$.

\begin{algorithm}
\caption{BCD Algorithm to Solve DLOT via UCOOT}
\label{alg:dlot}
\begin{algorithmic}[1]
\Require Representations $(\mathbf{h}_{\text{orig}}, \mathbf{p}_{\text{orig}})$, $(\mathbf{h}_{\text{pert}}, \mathbf{p}_{\text{pert}})$, interactions $\xi_{\text{orig}}, \xi_{\text{pert}}$, weights $\lambda_1, \lambda_2$, entropic regularization $\varepsilon$, iterations $T$
\Ensure The final transport plans $\pi^s$, $\pi^f$
\State Initialize $\pi^s$, $\pi^f$
\For{$t = 1, \dots, T$} % 使用 \For（大写 F）
    \State Compute cost matrix $C^s$ using current $\pi^f$
    \State Update $\pi^s$ via entropic Sinkhorn (see below)
    \State Rescale $\pi^s \leftarrow \sqrt{m(\pi^f)/m(\pi^s)} \cdot \pi^s$
    \State Compute cost matrix $C^f$ using updated $\pi^s$
    \State Update $\pi^f$ via entropic Sinkhorn
    \State Rescale $\pi^f \leftarrow \sqrt{m(\pi^s)/m(\pi^f)} \cdot \pi^f$
\EndFor % 使用 \EndFor（大写 F）
\end{algorithmic}
\end{algorithm}

% \begin{algorithm}
% \caption{BCD Algorithm to Solve DLOT via UCOOT}
% \label{alg:dlot}
% \begin{algorithmic}[1]
% \Require Representations $(\mathbf{h}_{\text{orig}}, \mathbf{p}_{\text{orig}})$, $(\mathbf{h}_{\text{pert}}, \mathbf{p}_{\text{pert}})$, interactions $\xi_{\text{orig}}, \xi_{\text{pert}}$, weights $\lambda_1, \lambda_2$, entropic regularization $\varepsilon$, iterations $T$
% \ENSURE The final transport plans $\pi^s$, $\pi^f$
% \STATE Initialize $\pi^s$, $\pi^f$
% \FOR{$t = 1, \dots, T$}
%     \STATE Compute cost matrix $C^s$ using current $\pi^f$
%     \STATE Update $\pi^s$ via entropic Sinkhorn (see below)
%     \STATE Rescale $\pi^s \leftarrow \sqrt{m(\pi^f)/m(\pi^s)} \cdot \pi^s$
%     \STATE Compute cost matrix $C^f$ using updated $\pi^s$
%     \STATE Update $\pi^f$ via entropic Sinkhorn
%     \STATE Rescale $\pi^f \leftarrow \sqrt{m(\pi^s)/m(\pi^f)} \cdot \pi^f$
% \ENDFOR
% \end{algorithmic}
% \end{algorithm}

\paragraph{Sinkhorn Update}

Each transport plan update in Algorithm~\ref{alg:dlot} solves an unbalanced optimal transport problem with KL divergence regularization. The generic form is:
\begin{equation}
\label{eq:sinkhorn_obj}
\min_{\pi \in \mathbb{R}_+^{n \times m}} \langle C, \pi \rangle + \lambda \, \mathrm{KL}(\pi \| r \otimes c) + \varepsilon \, \mathrm{KL}(\pi \| \mathbf{1}),
\end{equation}
where $C \in \mathbb{R}^{n \times m}$ is the cost matrix, $r \in \Delta^n$, $c \in \Delta^m$ are relaxed (non-normalized) marginal histograms (e.g., uniform), $\lambda$ controls the divergence penalty (mass conservation), $\varepsilon$ is the entropic regularization weight.

The solution is obtained via multiplicative Sinkhorn iterations. Define the Gibbs kernel $K = \exp(-C / \varepsilon)$. The update proceeds is that we first initialize $u \leftarrow \mathbf{1}_n$ and $v \leftarrow \mathbf{1}_m$, then we repeat for $T_{\text{inner}}$  iterations or until convergence:
\begin{equation}
    \quad u \leftarrow \left( \frac{r}{K v} \right)^{\lambda / (\lambda + \varepsilon)} \label{eq:sinkhorn_u},
\end{equation}
\begin{equation}
   \quad v \leftarrow \left( \frac{c}{K^\top u} \right)^{\lambda / (\lambda + \varepsilon)} \label{eq:sinkhorn_v}.
\end{equation}
Finally, the transport plan is reconstructed as:
\begin{equation}
\pi = \operatorname{diag}(u) \cdot K \cdot \operatorname{diag}(v).
\end{equation}

This iterative scheme solves the optimality conditions of Eq.~\eqref{eq:sinkhorn_obj}. The exponent $\lambda / (\lambda + \varepsilon)$ balances mass fidelity and entropy.

\paragraph{Numerical Stabilization}

To prevent underflow when $\varepsilon$ is small or $C$ has large entries, it is common to perform the iterations in the log-domain:
\begin{equation}
\begin{aligned}
\log u \leftarrow \frac{\lambda}{\lambda + \varepsilon} \left( \log r - \log(K v) \right), \\
\log v \leftarrow \frac{\lambda}{\lambda + \varepsilon} \left( \log c - \log(K^\top u) \right).
\end{aligned}
\end{equation}

\paragraph{Cost Matrix Computation}

In our DLOT context, the cost matrices $C^s$ and $C^f$ are recomputed at each BCD iteration using the other coupling:
\begin{align}
C^s_{ij} &= \sum_{k,l} \left| \xi_{\text{orig}}(h_i, p_k) - \xi_{\text{pert}}(h_j, p_l) \right|^p \cdot \pi^f_{kl}, \\
C^f_{kl} &= \sum_{i,j} \left| \xi_{\text{orig}}(h_i, p_k) - \xi_{\text{pert}}(h_j, p_l) \right|^p \cdot \pi^s_{ij}.
\end{align}

These are weighted average discrepancies between sample-feature interactions across domains, used to guide alignment.

\section{B \  Derivation of Constrained Reinforcement with Dynamic Barrier}

To dynamically adjust the penalty coefficient $\delta$ in the Lagrangian $\mathcal{L}(\psi, \delta)$, we formulate a constrained optimization subproblem that balances reward maximization and constraint satisfaction. This derivation extends the bi-level optimization framework to reinforcement learning with policy gradients. Below, we provide a step-by-step expansion of the mathematical reasoning.

% ========================
\subsection{Constrained Optimization Reformulation}
% ========================
The primal optimization problem aims to maximize the combined rewards $\mathcal{J}_{\text{dir}}$ and $\mathcal{J}_{\text{div}}$ while ensuring the stealth constraint $\mathcal{J}_{\text{dist}} \leq \rho_{st}$ is satisfied. Formally:
\begin{equation}
    \max_{\psi} \, \underbrace{\mathcal{J}_{\text{dir}}(\psi) + \mathcal{J}_{\text{div}}(\psi)}_{\text{Reward Objectives}} \quad \text{s.t.} \quad \underbrace{\mathcal{J}_{\text{dist}}(\psi) \leq \rho_{st}}_{\text{Stealth Constraint}}.
\end{equation}
To handle the constraint, we employ Lagrangian relaxation, which converts the constrained problem into a saddle-point problem. The Lagrangian $\mathcal{L}(\psi, \delta)$ introduces a penalty coefficient $\delta \geq 0$ to penalize constraint violations:
\begin{equation}
    \min_{\delta \geq 0} \max_{\psi} \, \mathcal{L}(\psi, \delta) = \mathcal{J}_{\text{dir}} + \mathcal{J}_{\text{div}} - \delta \left( \mathcal{J}_{\text{dist}} - \rho_{st} \right).
\end{equation}
Here, $\delta$ acts as a dual variable that adaptively scales the penalty based on the degree of constraint violation $\mathcal{J}_{\text{dist}} - \rho_{st}$. The min-max formulation allows simultaneous optimization of the policy $\psi$ (maximizing rewards) and the penalty $\delta$ (minimizing constraint violations).

% ========================
\subsection{Gradient Alignment Objective}
% ========================
A critical challenge arises when gradients of the reward objectives ($\nabla \mathcal{J}_{\text{dir}}, \nabla \mathcal{J}_{\text{div}}$) and constraint ($\nabla \mathcal{J}_{\text{dist}}$) conflict. To mitigate this, we design a secondary objective $\mathcal{J}(\delta)$ that enforces alignment between reward and constraint gradients. Specifically, $\mathcal{J}(\delta)$ minimizes the Euclidean distance between the combined reward gradient and the scaled constraint gradient, while also penalizing constraint violations:
\begin{equation}
    \mathcal{J}(\delta) = \underbrace{\frac{1}{2} \left\| \nabla_{\psi}(\mathcal{J}_{\text{dir}} + \mathcal{J}_{\text{div}}) - \delta \nabla_{\psi} \mathcal{J}_{\text{dist}} \right\|^2}_{\text{Gradient Alignment}} - \underbrace{\delta \left( \mathcal{J}_{\text{dist}} - \rho_{st} \right)}_{\text{Violation Penalty}}.
\end{equation}
The first term ensures that the direction of policy updates (driven by rewards) does not contradict the constraint gradient. The second term directly penalizes deviations from the stealth bound $\rho_{st}$.

% ========================
\subsection{Solve for Optimal $\delta$}
% ========================
To find the optimal $\delta$, we take the derivative of $\mathcal{J}(\delta)$ with respect to $\delta$ and set it to zero:
\begin{equation}
\begin{aligned}
    \frac{\partial \mathcal{J}}{\partial \delta} 
    = ( \nabla_{\psi} \mathcal{J}_{\text{dist}} )^\top ( \nabla_{\psi}(\mathcal{J}_{\text{dir}} + \mathcal{J}_{\text{div}}) - \\
    \delta \nabla_{\psi} \mathcal{J}_{\text{dist}} ) - (\mathcal{J}_{\text{dist}} - \rho_{st}) = 0.
\end{aligned}
\end{equation}
Rearranging terms yields a closed-form solution for $\delta$:
\begin{equation}
    \delta = \frac{ \overbrace{\left( \mathcal{J}_{\text{dist}} - \rho_{st} \right)}^{\text{Constraint Violation}} + \overbrace{\nabla_{\psi} \mathcal{J}_{\text{dist}}^\top \nabla_{\psi}(\mathcal{J}_{\text{dir}} + \mathcal{J}_{\text{div}})}^{\text{Gradient Conflict Term}} }{ \underbrace{\left\| \nabla_{\psi} \mathcal{J}_{\text{dist}} \right\|^2}_{\text{Normalization by Constraint Gradient Magnitude}} }.
\end{equation}
This expression reveals that $\delta$ increases when the constraint is violated ($\mathcal{J}_{\text{dist}} > \rho_{st}$) or when the reward and constraint gradients are aligned (positive inner product). Conversely, conflicting gradients reduce $\delta$ to prioritize reward maximization.

% ========================
\subsection{Policy Gradient Expansion}
% ========================
Using the policy gradient theorem, we express the gradients $\nabla_{\psi} \mathcal{J}_r$ ($r \in \{\text{dir}, \text{div}, \text{dist}\}$) as expectations over trajectories sampled from the policy $\pi_\psi$. For each reward or constraint term:
\begin{equation}
    \nabla_{\psi} \mathcal{J}_r = \mathbb{E}_{\tau \sim \pi_\psi} \left[ \sum_{t=0}^T \nabla_{\psi} \log \pi_{\psi}(a_t|s_t) \cdot A_{R_r}(s_t, a_t) \right],
\end{equation}
where $A_{R_r}(s_t, a_t)$ denotes the advantage function for reward $R_r$, estimating how much better an action $a_t$ is compared to the average at state $s_t$. Substituting these expectations into the expression for $\delta$ gives:
\begin{align}
    &\delta =
   \frac{\left( \mathcal{J}_{\text{dist}} - \rho_{st} \right) }{\left\| \mathbb{E}_\tau \left[ \sum_t \nabla_{\psi} \log \pi_{\psi} \cdot A_{\mathcal{J}_{\text{dist}}} \right] \right\|^2}+ \\ 
    &\frac{ 
         \mathbb{E}_\tau \left[ \sum_t \nabla_{\psi} \log \pi_{\psi} \cdot A_{\mathcal{J}_{\text{dist}}} \right]^\top \mathbb{E}_\tau \left[ \sum_t \nabla_{\psi} \log \pi_{\psi} \cdot (A_{\mathcal{J}_{\text{dir}}} + A_{\mathcal{J}_{\text{div}}}) \right] 
    }{ 
        \left\| \mathbb{E}_\tau \left[ \sum_t \nabla_{\psi} \log \pi_{\psi} \cdot A_{\mathcal{J}_{\text{dist}}} \right] \right\|^2 
    }.
\nonumber
\end{align}
This formulation explicitly connects $\delta$ to the advantage-weighted policy gradients, ensuring updates account for both immediate and long-term effects of actions.

% ========================
\subsection{Dynamic Barrier Regularization}
% ========================
To ensure numerical stability and avoid division by zero, we add a small constant $\kappa > 0$ to the denominator. Additionally, we enforce non-negativity of $\delta$ through the operator $[\,\cdot\,]_+ = \max(\cdot, 0)$, which projects negative values to zero:
\begin{equation}
    \delta = \left[ \frac{ 
    \mathcal{J}_{\text{dist}} - \rho_{st} - \nabla_{\psi} \mathcal{J}_{\text{dist}}^\top \nabla_{\psi}(\mathcal{J}_{\text{dir}} + \mathcal{J}_{\text{div}}) 
    }{ 
        \left\| \nabla_{\psi} \mathcal{J}_{\text{dist}} \right\|^2 + \kappa 
    } \right]_+.
% \nonumber
\end{equation}
The term $-\nabla_{\psi} \mathcal{J}_{\text{dist}}^\top \nabla_{\psi}(\mathcal{J}_{\text{dir}} + \mathcal{J}_{\text{div}})$ quantifies the conflict between reward and constraint gradients. If they oppose each other (negative inner product), $\delta$ decreases to prioritize reward maximization. Severe constraint violations ($\mathcal{J}_{\text{dist}} \gg \rho_{st}$) dominate the numerator, increasing $\delta$ to suppress detectable perturbations.
The policy gradients are derived as:
\begin{equation}
\small
\begin{aligned}
\nabla_{\psi}(\mathcal{J}_{\text{dir}} + \mathcal{J}_{\text{div}}) &= \mathbb{E}_\tau \left[ \sum_t \nabla_\psi \log \pi_\psi(a_t|s_t) \cdot (\hat{A}_{R_{dir}} + \hat{A}_{R_{div}}) \right], \\
\nabla_{\psi} \mathcal{J}_{\text{dist}}&= \mathbb{E}_\tau \left[ \sum_t \nabla_\psi \log \pi_\psi(a_t|s_t) \cdot \hat{A}_{R_{dist}}(s_t, a_t) \right].
\end{aligned}
% \nonumber
\end{equation}

% ========================
% \subsection*{Interpretation}
% ========================
The term $\mathcal{J}_{\text{dist}} - \rho_{st}$ directly measures the severity of stealth constraint violation. The inner product term evaluates the alignment between constraint gradients ($\nabla \mathcal{J}_{\text{dist}}$) and combined reward gradients ($\nabla \mathcal{J}_{\text{dir}} + \nabla \mathcal{J}_{\text{div}}$). Negative alignment (conflict) reduces $\delta$, allowing greater emphasis on rewards.
Normalizes the penalty by the squared magnitude of constraint gradients, stabilizing updates across varying gradient scales.
The constant $\kappa$ prevents division by near-zero values, ensuring numerical robustness.
When constraints are satisfied ($\mathcal{J}_{\text{dist}} \leq \rho_{st}$) and gradients align, $\delta$ remains small, prioritizing attack efficacy.
Severe violations or aligned gradients increase $\delta$, enforcing stealthiness without sacrificing convergence stability.

This derivation rigorously extends the bi-level optimization framework to policy gradient RL, enabling adaptive trade-offs between attack efficacy and stealth. In practice, expectations are approximated via Monte Carlo sampling over trajectories generated by the current policy $\pi_\psi$. The dynamic barrier mechanism ensures constraints are softly enforced throughout training, adapting to evolving policy and environmental dynamics.

\begin{algorithm}[ht]
\caption{The proposed framework CREAT}
\label{alg:creat}
\begin{algorithmic}[1]
\Require 
Sequential recommender $\mathbf{\Phi}_{\theta}$, 
Training sequences $\mathcal{S} \subseteq \mathcal{D}$, 
Target item $v^*$, 
Perturbation budget $K$
\Ensure Optimized perturbation masker $\mathcal{M}_{\psi}$

\State Initialize perturbation masker $\mathcal{M}_{\psi}$ with parameters $\psi$
\State Initialize group-shared experience replay buffer $\mathcal{B}$

\State{\textit{Stage 1: The Localization Stage}}
\For{each sequence $\mathbf{s} \in \mathcal{S}$}
    \State $\mathbf{s}'^{(0)} \gets \mathbf{s}$, $T^{(0)} \gets \emptyset$
    \For{step $i=1$ to $K$}
        \State Generate mask $\mathbf{m}^{(i)} \sim \mathcal{M}_{\psi}(\mathbf{s}'^{(i-1)})$
        \State Construct perturbed sequence: \\ \quad \quad  \quad 
        $\mathbf{s}'^{(i)} \gets \mathbf{s}'^{(i-1)} \odot (1-\mathbf{m}^{(i)}) + v^* \cdot \mathbf{m}^{(i)}$
        \State $T^{(i)} \gets T^{(i-1)} \cup \{t_j | \mathbf{m}_t^{(i)} = 1\}$
        \State Compute pattern inversion reward: \\ \quad \quad  \quad $R_{\text{inv}}^{(i)} = R_{\text{dir}}^{(i)} + R_{\text{div}}^{(i)}$
        \State Store transition $(\mathbf{s}'^{(i-1)}, \mathbf{m}^{(i)}, R_{\text{inv}}^{(i)}, \mathbf{s}'^{(i)})$ in $\mathcal{B}$
    \EndFor
\EndFor
\State Update $\psi$ via policy gradient: $\psi \gets \psi + \eta \nabla_{\psi} \mathcal{J}_{\text{inv}}$ 

\State{\textit{Stage 2: The Constrained Stage}}
\While{not converged}
    \State Sample group of $G$ trajectories from $\mathcal{B}$
    \For{each trajectory $(\mathbf{s}'^{(0)}, \ldots, \mathbf{s}'^{(K)})$}
        \For{each step $i$}
            \State Compute distribution consistency reward: \\ \quad \quad \quad \quad \quad 
            $R_{\text{dist}}^{(i)} = -\texttt{DLOT}\left(\mathbb{X}_{\text{orig}}, \mathbb{X}_{\text{pert}}^{(i)}\right)$
        \EndFor
    \EndFor
    \State Compute group-wise baseline $\mu_G$ and $\sigma_G$
    \State Set dynamic stealth threshold: \\ \quad \quad  $\rho_{st} = \mu_G^{\text{dist}} + \lambda_{st} \cdot \sigma_G^{\text{dist}}$
    \State Calculate dynamic barrier coefficient:\\ \quad \quad 
    $\delta = \left[\frac{\mathcal{J}_{\text{dist}} - \rho_{st} - \nabla_{\psi}\mathcal{J}_{\text{dist}}^{\top}\nabla_{\psi}(\mathcal{J}_{\text{dir}} + \mathcal{J}_{\text{div}})}{\|\nabla_{\psi}\mathcal{J}_{\text{dist}}\|^2 + \kappa}\right]_+$
    \State Compute normalized advantage estimates: \\  \quad \quad  $\hat{A}_{i,t} = \frac{r_{i,t} - \mu_{G,t}}{\sigma_{G,t} + \epsilon}$
    \State Update policy with constrained gradient: \\ \quad \quad 
    $\nabla_{\psi}\mathcal{L} = \nabla_{\psi}\mathcal{J}_{\text{dir}} + \nabla_{\psi}\mathcal{J}_{\text{div}} - \delta \cdot \nabla_{\psi}\mathcal{J}_{\text{dist}}$
    \State \quad $\psi \gets \psi + \eta \nabla_{\psi}\mathcal{L}$
\EndWhile
\end{algorithmic}
\end{algorithm}

\section{C \  Algorithm Summary}
The algorithm of our proposed model \modelname~is outlined in a two-stage optimization process in Algorithm \ref{alg:creat}.
From lines 1-12, we introduce the Localization Stage, which initializes the perturbation masker and processes each training sequence. This phase involves generating perturbation masks and constructing perturbed sequences step by step while calculating the pattern inversion rewards. The transition at each step is stored for later analysis.
From Lines 13-27, we describe the Constrained Fine-tuning Stage, focusing on fine-tuning the model through reinforcement learning. Here, the policy is updated using a constrained gradient to ensure both attack intensity, i.e., maximizing pattern inversion, and stealthiness i.e., minimizing distribution shifts. Group-wise baselines are calculated, and dynamic barriers adjust the penalty term to ensure that perturbations remain stealthy yet effective. This dual-level optimization ultimately converges to a policy capable of performing subtle, high-impact profile pollution attacks against sequential recommendation.

\section{D \  Datasets and Implementation Details}

\subsection{Dataset Details}
We use three real-world recommendation datasets to evaluate the proposed methods, i.e. MovieLens-1M (ML-1M), MovieLens-20M (ML-20M), and Amazon Beauty. We employ 5-core versions of the datasets and preprocess them according to \cite{bert4reccc}. The last two items in each sequence are reserved for validation and testing, with the rest assigned to training. The details of the pre-processed datasets are shown in Table \ref{tab:dataset}.

\begin{table}[ht]
\small
\centering
\begin{tabular}{cccccc}
    \toprule
    \textbf{Datasets} & \textbf{\#User} & \textbf{\#Item} & \textbf{Avg.len} & \textbf{Sparsity} \\
    \midrule
    ML-1M  & 6,040 & 3,416 & 165 & 95.2\%\\
    \midrule
    Beauty & 22,363  & 12,101 & 9 & 99.9\%\\
    \midrule
    ML-20M & 138,493 & 18,345 & 144 & 99.2\%\\
    \bottomrule
\end{tabular}
\caption{Statistics on Datasets.}
\label{tab:dataset}
\vspace{-0.1 in}
\end{table}

\subsection{Implementation Details}
For fair comparison, we follow the implementation in previous work when implementing backbone SR models and all baselines.
The backbone models are uniformly trained using the Adam optimizer with a learning rate of 0.001, weight decay of 0.01, and a fixed batch size of 64.
In the reinforcement learning training, the Adam optimizer is employed with a learning rate of 1e-5.
For the group-relative policy optimization, the group size is set to 32, and the number of policy updates per round is 20.
Following previous works\cite{blackbox,defending}, we set the maximum sequence length to 200 for the ML-1M and ML-20M datasets and 50 for the Beauty dataset.
In the baseline experiments, the maximum number of replacements is set to 2 for ML-1M and ML-20M, and 1 for Beauty, with the proportion of manipulable sequence subsets controlled by the attacker fixed at 10\%.
To evaluate the effectiveness of targeted attacks, we focus on metrics tailored to a specific targeted item. Hit Rate (HR@K) measures if the target item appears in the top-K recommendations:
\begin{equation}
    \text{HR@K} = \frac{\sum_{u \in \mathcal{U}} \mathbb{I}(\text{rank}_u^{\text{target}} \leq K)}{|\mathcal{U}|},
\end{equation}
where $\mathbb{I}$ is an indicator function, $\mathcal{U}$ is the user set, and $\text{rank}_u^{\text{target}}$ is the rank of the targeted item for user $u$. Normalized Discounted Cumulative Gain (NDCG@K) evaluates the ranking position of the target item: 
\begin{equation}
    \text{NDCG@K} = \frac{1}{|\mathcal{U}|} \sum_{u \in \mathcal{U}} \frac{\mathbb{I}(\text{rank}_u^{\text{target}} \leq K) \cdot \frac{1}{\log_2(\text{rank}_u^{\text{target}} + 1)}}{\text{IDCG@K}_u}.
\end{equation}
Here, $\text{IDCG@K}_u$ is the ideal DCG@K (computed if the target item were ranked first). Mean Reciprocal Rank (MRR) assesses the average reciprocal rank of the targeted item: 
\begin{equation}
    \text{MRR} = \frac{1}{|\mathcal{U}|} \sum_{u \in \mathcal{U}} \frac{1}{\text{rank}_u^{\text{target}}}.
\end{equation}
We conduct our experiment on a Ubuntu OS with version 22.04.4 LTS, which contains 4 NVIDIA RTX 3090 GPUs, 2 64-bit 12-core Intel(R) Xeon(R) Silver 4116 CPUs @ 2.10GHz and 503GB of RAM.

\section{E \  Attack Performance Under Defense}
To evaluate the stealthiness of \modelname~, we employ the defense method ADVTrain \cite{defending}, which is specifically designed for profile pollution attacks.
We utilize ADVTrain to defend against three attack SOTAs and our attack \modelname~on two SR backbones.
From the results in Table \ref{tab:defense}, we can conclude:
1) \modelname~causes less disruption of attack performance than other attacks (measured by relative degradation: \(\frac{P_{\text{undef}} - P_{\text{def}}}{P_{\text{undef}}}\), where \(P_{\text{undef}}\) and \(P_{\text{def}}\) represent attack performance on undefended and defended models, respectively), indicating the superiority of \modelname~in evading the ADVTrain defense.
This further validates the efficacy of the constraint reward on the perturbed sequence distribution in \modelname, which minimizes the perturbation magnitude while achieving similar attack effectiveness, thereby weakening the defense method's ability to detect and counter the perturbed sequences.
2) \modelname~achieves powerful attack performance on both backbones with or without defending, demonstrating its robustness across different scenarios.

\begin{table}[ht]
\footnotesize
\centering
% \vspace{-0.1in}
\resizebox{1.0\linewidth}{!}{
\begin{tabular}{c|cc|cc}
\toprule
 \multirow{2}{*}{\textbf{Attack}}
& \multicolumn{2}{c|}{\textbf{NARM}}
& \multicolumn{2}{c}{\textbf{Bert4Rec}}\\
 % \cmidrule(lr){1-1} \cmidrule(lr){2-4} \cmidrule(lr){5-7}
\cmidrule(lr){2-3} \cmidrule(lr){4-5}

% \multirow{2}{*}{ER@10 (AZ, IJCAI)} & \multicolumn{2}{c}{$\mu=0.1\%$} & \multicolumn{2}{c}{$\mu=1.0\%$} & \multicolumn{2}{c}{$\mu=10\%$} & \multicolumn{2}{c}{$\mu=0.1\%$} & \multicolumn{2}{c}{$\mu=1.0\%$} & \multicolumn{2}{c}{$\mu=10\%$} \\
& \multicolumn{1}{c|}{H@10} & \multicolumn{1}{c|}{N@10} & \multicolumn{1}{c|}{H@10} & \multicolumn{1}{c}{N@10}\\
% \cmidrule(lr){2-3} \cmidrule(lr){4-5}\cmidrule(lr){6-7} \cmidrule(lr){8-9} \cmidrule(lr){10-11} \cmidrule(lr){12-13}
% & HR@10 & NDCG@10 & HR@10 & NDCG@10 & HR@10 & NDCG@10 & HR@10 & NDCG@10 & HR@10 & NDCG@10 & HR@10 & NDCG@10\\

 \midrule
% ML-1M
SimAlter & 0.0257(↓51.14\%) & 0.0103(↓51.42\%) & 0.0201(↓55.73\%) & 0.0099(↓48.97\%)\\
Replace & 0.0464(↓38.95\%) & 0.0177(↓44.39\%) & 0.0309(↓51.49\%) & 0.0163(↓52.19\%)\\
SSLAttack & 0.0417(↓44.69\%) & 0.0165(↓48.02\%) & 0.0383(↓44.81\%) & 0.0201(↓45.97\%)\\
\rowcolor{gray!18} \modelname & \textbf{0.0825(↓27.12\%)}  & \textbf{0.0379(↓30.84\%)}  & \textbf{0.0787(↓26.45\%)}  & \textbf{0.0351(↓29.52\%)} \\
\bottomrule

\end{tabular}
}
% \vspace{-0.1 in}
\caption{Attack performance under defenses.}
\label{tab:defense}
\vspace{-0.15 in}
\end{table}

\section{F \  Discussion on Potential Positive and Negative Societal Impacts}

The proposed \modelname~framework for profile pollution attacks against sequential recommendation systems presents several potential societal impacts that warrant careful consideration. On the positive side, this work contributes to the broader understanding of adversarial vulnerabilities in recommendation systems, a critical step toward building more secure and trustworthy AI systems. By exposing the risks of stealthy perturbations targeting fine-grained sequential patterns, this research motivates the development of robust defense mechanisms, ultimately enhancing the resilience of recommendation algorithms against malicious manipulation. Furthermore, the bi-level optimization framework and constrained reinforcement learning paradigm introduced in \modelname~may inspire novel methodologies for adversarial robustness research beyond recommendation systems, such as in fraud detection or anomaly identification, where balancing efficacy and stealthiness is paramount. Conversely, the attack methodology itself could be misused by malicious actors to artificially inflate the visibility of specific items (e.g., misinformation, counterfeit products, or biased content) while evading detection, undermining user trust and platform integrity. To mitigate such risks, the authors explicitly emphasize the necessity of future work on adaptive defense strategies and advocate for responsible disclosure through controlled code release. By proactively addressing these dual-use implications, this work aligns with ethical AI research practices, advancing security knowledge while highlighting the urgency of safeguarding recommendation ecosystems.

\end{document}